\documentclass[11pt]{article}
\usepackage{EMNLP2022}
\usepackage{times}
\usepackage{subfig}
\usepackage{graphics}
\usepackage{graphicx}
\usepackage{latexsym}
\usepackage{rotating}
\usepackage{adjustbox}
\usepackage{multirow}

\usepackage{amsmath}
\usepackage{hyperref}
\usepackage{algpseudocode}
\usepackage{makecell}
\usepackage[linesnumbered,ruled,vlined]{algorithm2e}
\usepackage{microtype}
\SetKwInput{KwInput}{Input}                
\SetKwInput{KwOutput}{Output}              

\makeatletter
\newcommand{\printfnsymbol}[1]{%
  \textsuperscript{\@fnsymbol{#1}}%
}
\title{T-STAR:  \underline{T}ruthful \underline{S}tyle \underline{T}ransfer using \underline{A}MR Graph as  Intermediate \underline{R}epresentation}

\author{Anubhav Jangra\thanks{\, denotes equal contribution}\quad Preksha Nema\printfnsymbol{1}\quad Aravindan Raghuveer \\
  Google Research, India\\
  \texttt{\{anubhavjangra,preksh,araghuveer\}@google.com}  \\}

\date{}

\begin{document}
\maketitle
\begin{abstract}
Unavailability of parallel corpora for training text style transfer  (TST) models is a very challenging yet common scenario. Also, TST models implicitly need to preserve the content  while transforming a source sentence into the target style.  To tackle these problems,  an intermediate representation is often constructed that is devoid of style while still preserving the meaning of the source sentence.  In this work, we study the usefulness of Abstract Meaning Representation (AMR) graph as the intermediate style agnostic representation. We posit that semantic notations like AMR are a natural choice for an intermediate representation. Hence, we propose  \textbf{T-STAR}: a model comprising of two components, text-to-AMR encoder and a AMR-to-text decoder. We propose several modeling improvements to enhance the style agnosticity of the generated AMR. To the best of our knowledge, T-STAR is the first work that uses AMR as an intermediate representation for TST.  With thorough experimental evaluation we show T-STAR significantly outperforms state of the art techniques by achieving on an average $15.2$\% higher content preservation with negligible loss ($\sim$3\%) in style accuracy. Through detailed human evaluation with $90,000$ ratings, we also show that T-STAR has upto $50$\% lesser hallucinations compared to state of the art TST models.  

\end{abstract}
\section{Introduction}

A well accepted definition of {\em style} refers to the manner (via linguistic elements like word choices, syntactic structures, metaphors)  in which semantics of a sentence are expressed~\cite{mcdonald1985computational,styletransfer-survey}.
Text Style Transfer (TST) is the task of rephrasing an input sentence to contain specific stylistic properties without altering the meaning of the sentence~\cite{styletransfer-bt}. We refer the reader to  \citet{styletransfer-survey} for a detailed survey of  approaches towards TST problem formulation, metrics and models. In the practical scenario, that we consider in this paper, where a large corpus of parallel data is   not available~\cite{polite-dialoggeneration, online-text-debiasing, stylistic-mt}, two family of approaches have been proposed in the literature~\cite{styletransfer-survey}.
\textbf{1. Disentanglement:} Content and Style are disentangled in a latent space and only the style information is varied to transform the sentence.
 \textbf{2. Prototype Editing:} Style bearing words in the source sentence are replaced with those corresponding to the target style. The sentence may be further re-arranged for fluency and naturalness.
Both the above approaches have drawbacks in the way the {\em style agnostic} intermediate representation is constructed, described as follows.   First, in the disentangling approaches, it is not easy to verify the efficacy of separation between style and content. Recent approaches ~\cite{multiple-attribute, samanta-etal-2021-hierarchical} even state that as content and style are so subtly entangled in text, it is difficult to disentangle both of them in a latent space.  Consequently,  this affects the model's interpretability in that it is hard to attribute an effect we see in the output to the latent intermediate vector or the style vector.     Second, with prototype editing approaches for lingustic styles (such as author, formality), where the content and style are tightly coupled it is not feasible to segregate style and content carrying words (word examples: {\em cometh, thou}). Furthermore, style-marker detection is a non-trivial NLP task and needs to be addressed for every new style that is added to the system causing scalabilty concerns ~\cite{styletransfer-survey}.
\begin{figure}
    \includegraphics[width=\columnwidth]{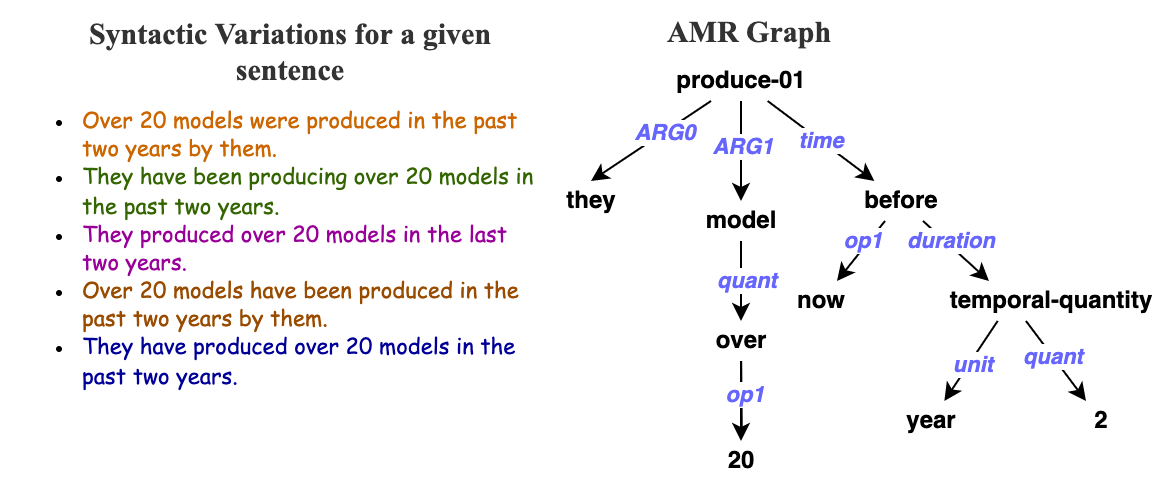}
    \caption{Different syntactic variations, leads to the same AMR as they all are similar in meaning.}
    \label{fig:amr_graph}
\end{figure}
In this paper, we propose \textbf{T-STAR} (\textbf{T}ruthful \textbf{S}tyle \textbf{T}ransfer using \textbf{A}MR Graph as  Intermediate \textbf{R}epresentation)     that uses  a symbolic semantic graph notation called Abstract Meaning Representation (AMR) as the style agnostic intermediate stage.  AMR  \cite{banarescu-etal-2013-abstract} is designed to capture semantics  of a given sentence in all entirety while abstracting away the syntactic variations, inflections and function words. In other words, two sentences with the same meaning but written in very different styles will have a very similar AMR if not exactly the same (See Figure~\ref{fig:amr_graph}). 

This addresses the shortcomings with {\em Disentanglement} and {\em Prototype Editing} approaches. First, AMR being a representation with well-defined  semantics, we can inspect, interpret and measure quality of the intermediate representation and the provenance of knowledge transfer between source and target sentence. Second, 
AMR being a well known standard has high quality, robust reference implementations, especially for head languages (\textit{e.g.}, English).   
Our contributions are as follows:

\noindent{\bf 1.}  We propose a novel TST approach with AMRs, an interpretable, symbolic intermediate representation, to achieve better content preservation. To this end, we enhance AMR parsing techniques to better suit the TST task.  To the best of our knowledge, we are the first work to use AMR representations for the style transfer task (Sections~\ref{sec:whyAMR},~\ref{sec:solution}).  
   
 \noindent {\bf 2.} Through novel experimentation, we show that an AMR, as a style agnostic intermediate representation, has better content preservation and less style information of the given source sentence compared to competitive baselines.(Sections~\ref{sec:experimental_setup},~\ref{sec:amr-robustness})

  \noindent {\bf 3.}  {On multiple datasets we show  T-STAR is able to beat competitive baselines by producing sentences with significantly
  higher content preservation  with similar style transfer scores.   (Section~\ref{sec:tst_perf})}
 
 \noindent {\bf 4.} {With thorough human evaluations spanning 90,000 ratings we show T-STAR has $\sim70$\% better content preservation compared to state of art baseline with $50$\% lesser hallucinations. (Section~\ref{sec:human_eval})}

\begin{figure*}[ht]
  \subfloat[MRPC]{\includegraphics[width=0.3\linewidth]{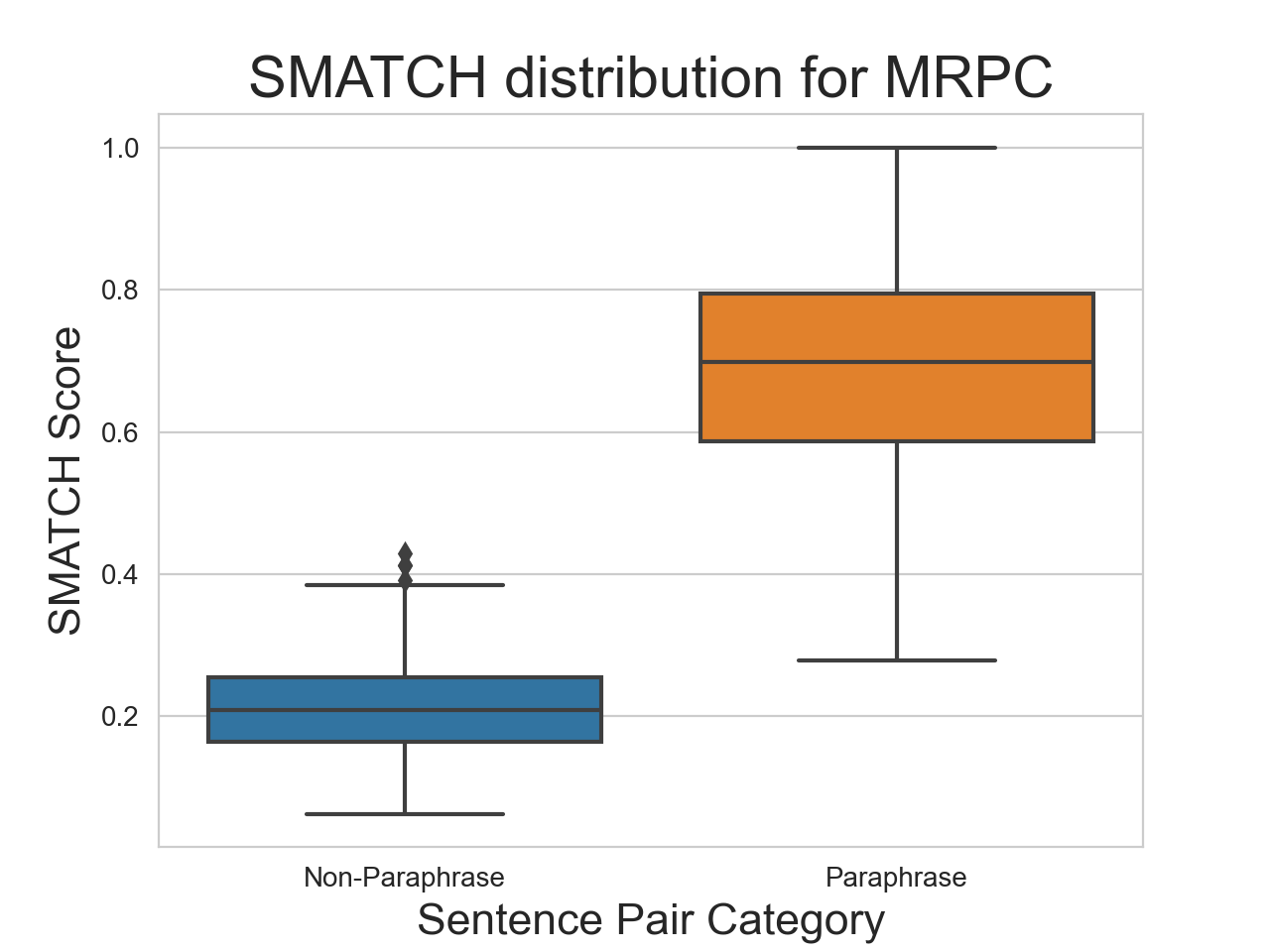}}
    \quad
    \subfloat[QQP]{\includegraphics[width=0.3\linewidth]{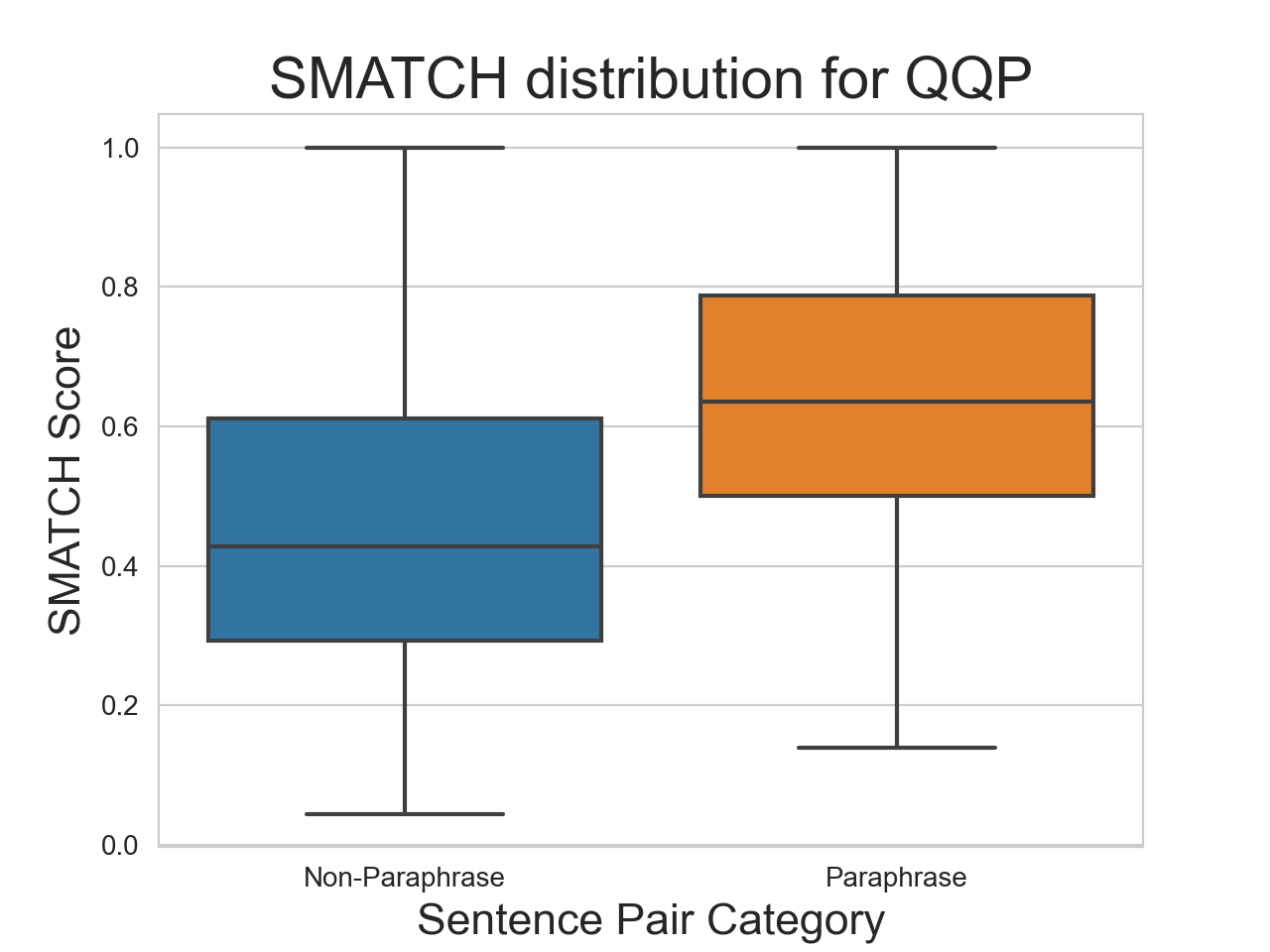}}
    \quad
         \subfloat[STS]{\includegraphics[width=0.3\linewidth]{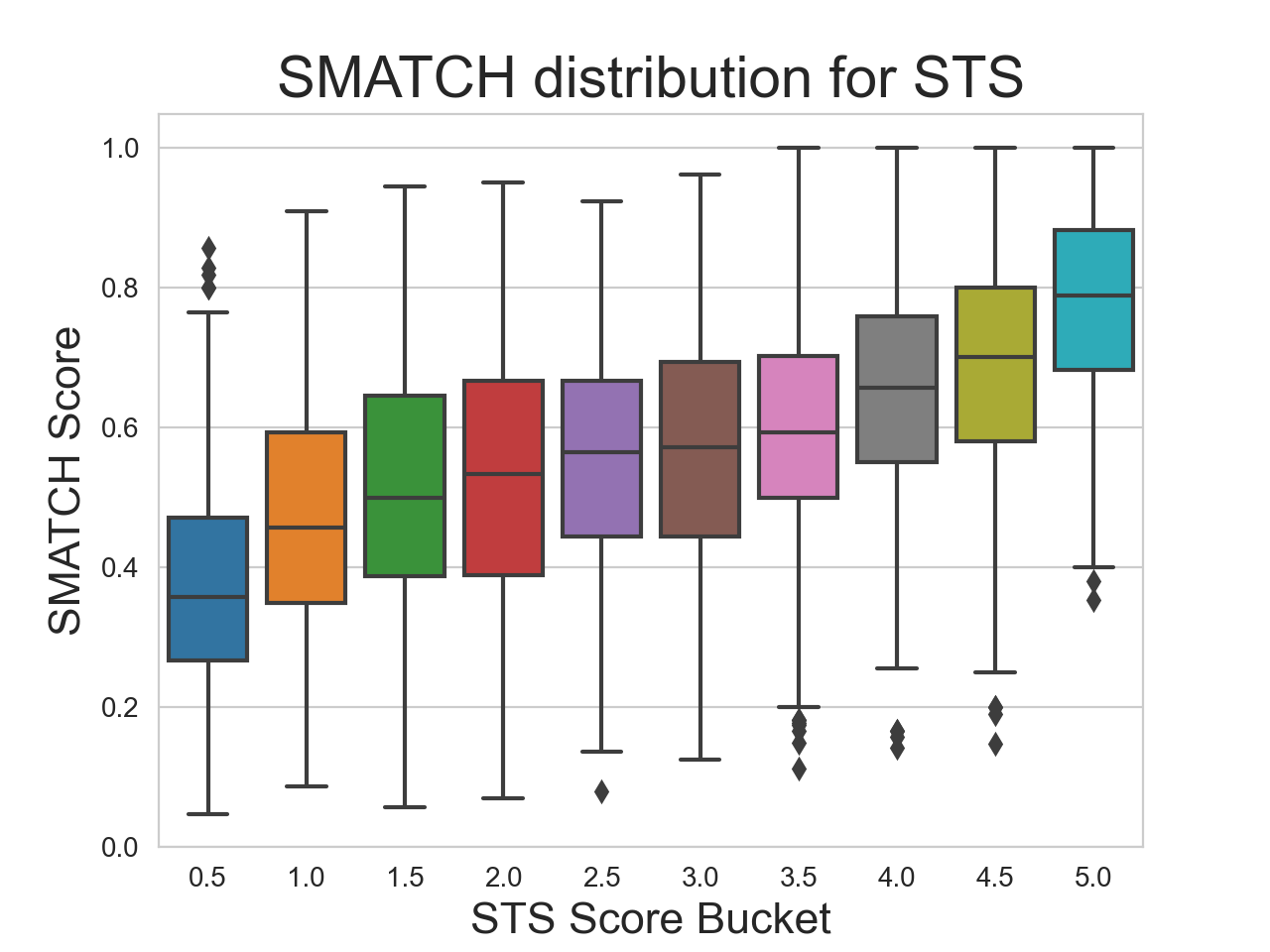}}
     \caption{AMR comparison for MRPC, QQP, and STS datasets}
     \label{amr-seq2seq}
\end{figure*}

\section{Related Work}
\subsection{Unsupervised Text Style Transfer} 
TST systems broadly use two family of approaches: disentanglement \cite{shen2017style} and prototype-editing \cite{styletransfer-survey}. Prior works \cite{neuralcontrolledtext,john-etal-2019-disentangled,adversarial-disentanglment,adversarial-st,content-preserving} disentangle the content and style information in latent space using style-based classifier or adversarial learning. Prototype-editing based approaches \cite{delete-retrieve,prototypeediting-madaan, prototypeediting-sudhakar} are used to gain better controllability and interpretability. 

Recently, some works propose to jointly optimize for content and style information, to overcome the limitations of explicitly disentangling the style and content information. \citet{yamshchikov2019decomposing} illustrates that architectures with higher quality of information decomposition perform better on style transfer tasks. \citet{multiple-attribute} argues that it is often easy to fool style discriminators without explicitly disentangling content and style information, which may lead to low content preservation \cite{xu2018unpaired}. Instead, \citet{multiple-attribute,content-preserving} use back-translation to optimize for content preservation. \citet{reinforce-st, st-reward-semanticsimilarity} use reinforcement learning framework with explicit rewards designed for content preservation. \citet{fastgradientapproach} pushes the entangled latent representation in the targeted-style direction using style discriminators. \citet{samanta-etal-2019-improved} uses normalizing flow to infuse the content and style information back before passing it to the decoder. 

\citet{cheng2020contextual} proposes a context aware text style transfer framework using two separate encoders for input sentence and the context. Similar to our work, \citet{divpara} also generates an interpretable intermediate representation. The authors first paraphrase the given source to first convert it to a destylized version before passing it to the targeted style-specific decoder. Complementary to our work, \cite{10.1145/3448733} uses syntactic graphs (dependency tree) as an intermediate representation for attribute transfer to retain the linguistic structure. On the other hand, we focus on retaining the semantics using AMR graphs as intermediate representation and modifying the linguistic structure (authorship style). 

\subsection{Text $\Longleftrightarrow$ AMR}
 In order to improve the parsing performance for AMRs, neural models are receving increasing attention. Neural AMR parsers can  be divided into following categories: i) sequence-to-sequence based AMR-parsers \cite{xu2020improving}, ii) sequence-to-graph based AMR parsers \cite{zhang2019amr}, where the graph is incrementally built by spanning one node at a time.
A more detailed survey on related works can be found in Appendix \ref{app:extendrw}.

\section{AMR as an Intermediate Representation} \label{sec:whyAMR}

Abstract Meaning Representations (AMRs) is a semantic formalism construct 
that abstracts away the syntactic information and only preserves the semantic meaning, in a rooted, directed and acyclic graph.  In Figure \ref{fig:amr_graph}, we present certain syntactic variations (changing the voice, and tense) for a sentence without altering   meaning. All variations result in the same AMR graph. The nodes in the AMR graphs (\textit{``produce-01'', ``they'', ``before'', etc.}) are concepts (entities/predicates) that are canonicalized and mapped to semantic role annotations present in Propbank framesets\footnote{\url{https://propbank.github.io/}}. The edges (\textit{``ARG0'', ``time'', ``duration'', etc.}) are then relations between these concepts. In other words, AMRs aim at decoupling ``\textit{what to say}'' from ``\textit{how to say}'' in an interpretrable way. We posit that this could be beneficial for text style transfer, where the goal is  to alter the ``\textit{how to say}'' aspect while preserving ``\textit{what to say}''.


Recently, semantic meaning representation property of AMRs has been shown to be useful in other generation tasks.  
In abstractive summarization, \citet{liao-etal-2018-abstract} uses AMR as an intermediate representation to first obtain a summary AMR graph from document and then generate a summary from it. \citet{hardy-vlachos-2018-guided} use AMRs to cover for the lack of explicit semantic modeling in sequence-to-sequence models. In machine translation, \citet{amr-nmt} adopted AMRs to enforce meaning preservation while translating from English to German. For Paraphrase Generation, \citet{cai-etal-2021-revisiting} found that using AMRs as intermediate representation reduces the semantic drift.
Moreover, incorporating symbolic representations as an intermediate representation provides a way to effectively understand the reasons behind a model's shortcomings. We utilize this advantage to analyse the weaknesses of T-STAR in Section~\ref{sec:error_analysis}. 

In order to demonstrate the semantic meaning preservation property of AMRs, we design an experiment using 
three publicly available paraphrase benchmarks, \textit{i.e.}, MRPC \cite{mrpc-dataset}, QQP\footnote{\hyperlink{https://www.quora.com/profile/Ricky-Riche-2/First-Quora-Dataset-Release-Question-Pairs}{https://www.quora.com/profile/Ricky-Riche-2/First-Quora-Dataset-Release-Question-Pairs}}, and STS.
MRPC and QQP are sentence pair datasets with each  pair labeled {\em yes} if they are paraphrases of each other {\em no} otherwise.
STS dataset assigns a scores from 0 (not similar) to 5 (exactly similar) to a sentence pair. 
We hypothesize that if AMRs are indeed semantic meaning preserving, two   sentences with similar meaning should have highly similar AMRs. 
To measure the similarity between two AMRs, we use the \textsc{Smatch} score \cite{smatch} that calculates the number of overlapping triplets between two AMRs.   We use an off-the-shelf AMR parser  \footnote{\hyperlink{amrlib}{https://amrlib.readthedocs.io/en/latest/}} to generate AMR given a sentence. We plot the distribution of the \textsc{Smatch} scores for MRPC, QQP and STS datasets in Figure \ref{amr-seq2seq}. For MRPC, we infer that the \textsc{Smatch} scores for paraphrases are significantly higher than the \textsc{Smatch} scores for non-paraphrases. Similarly, for QQP, quartile distribution of \textsc{Smatch} scores for paraphrases is higher in comparison to non-paraphrases. For STS dataset, we observe a gradual increase in quartile distribution of \textsc{Smatch} scores as we move towards more similar sentences.

The  experiments above corroborate the claim that AMRs can preserve meaning under lexical variations like paraphrasing, tense and voice changes. Recent research discussed earlier, have successfully used this property to show task improvements. Building on the above qualitative, quantitative and prior research evidence, we further explore the applicability of AMR for the TST task. 

\begin{table}[t]
      \resizebox{1\columnwidth}{!}{
      \begin{tabular}{ll}
      \hline\hline
         \textbf{Vanilla T-STAR Encoder}  &  \textbf{T-STAR Encoder} \\ \hline
         \multicolumn{2}{l}{\textit{\textbf{Sentence: To make us feel existence, and to shew}}} \\
          \makecell[l]{\tt{(a/and}\\
\tt{\quad :op1(m/make-02}\\
\tt{\quad \quad :ARG1(f/feel-01}\\
\tt{\quad\quad\quad :ARG0(w/we)}\\
\tt{\quad\quad\quad :ARG1(e/exist-01} \\
\tt{\quad\quad\quad\quad :ARG1 w)))}\\
\tt{\quad :op2 \textcolor{red}{(s / shew-01}}\\
\tt{\quad\quad:ARG0 w))}} & \makecell[l]{\tt{(h/have-purpose-91}\\
\tt{\quad:ARG2(a/and}\\
\tt{\quad\quad:op1(m/make-02}\\
\tt{\quad\quad\quad:ARG1(f/feel-01}\\
\tt{\quad\quad\quad\quad:ARG0(w/we)}\\
\tt{\quad\quad\quad\quad:ARG1(e/exist-01}\\
\tt{\quad\quad\quad\quad\quad:ARG1 w)))}\\
\tt{\quad\quad:op2\textcolor{blue}{(s/show-01}}\\
\tt{\quad\quad\quad:ARG0 w}\\
\tt{\quad\quad\quad:ARG1 e)))}}
         \\
         \hline
         \multicolumn{2}{l}{\textit{\textbf{Sentence: But trust not this; too easy Youth, beware!}}} \\
         \makecell[l]{\tt{(m/multi-sentence}\\
    \tt{\quad:snt1(c/contrast-01}\\
\tt{\quad\quad:ARG2(t/trust-01}\\
\tt{\quad\quad\quad:polarity -}\\
\tt{\quad\quad\quad:mode imperative}\\
\tt{\quad\quad\quad:ARG0(y/you)}\\
\tt{\quad\quad\quad:ARG1(t2/this)))}\\
\tt{\quad:snt2(b/beware-01}\\
\tt{\quad\quad:mode imperative}\\
\tt{\quad\quad:ARG0(y2/youth}\\
\tt{\quad\quad\quad:ARG1-of(e/easy-05}\\
\tt{\quad\quad\quad\quad:ARG2-of(h/have-degree-91}\\
\tt{\quad\quad\quad\quad\quad:ARG1 y2}\\
\tt{\quad\quad\quad\quad\quad:ARG3 (t3/too))))))}} & \makecell[l]{\tt{(c/contrast-01} \\
\tt{\quad:ARG2 (a/and}\\
\tt{\quad\quad:op1 (t/trust-02}\\
\tt{\quad\quad\quad:polarity -}\\
\tt{\quad\quad\quad:mode imperative}\\
\tt{\quad\quad\quad:ARG0 (y/you}\\
\tt{\quad\quad\quad\quad:mod (y2/youth))}\\
\tt{\quad\quad\quad:ARG1 (t2/this))}\\
\tt{\quad\quad:op2 (h/have-degree-91}\\
\tt{\quad\quad\quad:ARG1 t2}\\
\tt{\quad\quad\quad:ARG2 (e/easy-05}\\
\tt{\quad\quad\quad\quad:ARG1 t2)}\\
\tt{\quad\quad\quad:ARG3 (t3/too))))}} \\
           \hline\hline
      \end{tabular}}
      \caption{Comparison between AMRs from vanilla T-STAR Encoder and T-STAR Encoder. T-STAR Encoder generates better AMRs for stylized sentences. }
      \label{tab:amr-differences}
  \end{table}

\begin{table*}
\resizebox{1\textwidth}{!}{
\begin{tabular} {|p{2cm}|p{3cm}|p{12cm}|p{6cm}|}
\hline
\textbf{Dimension} & \textbf{Metric} & \textbf{Description} & \textbf{Metric used in related works} \\ 
\hline
AMR similarity & SMATCH$\uparrow$ & SMATCH \cite{smatch} measures degree of overlap between AMR graphs of \emph{S,T}. The score is computed based on triplet (edge) overlap by finding a variable node mapping that maximizes the count of matching triplets.  & Used extensively to measure similarity between two AMRs, across AMR-parsing literature.\\ \hline
Lexical Diveristy & Self-BLEU $\downarrow$ & BLEU-4 \cite{bleu} between \emph{S,T} & \cite{neuralcontrolledtext,delete-retrieve, content-preserving,xu2018unpaired}\\ \hline
\multirow{2}{*}{\makecell[l]{Content\\Preservation \\ (C.P.)}} & WMD $\downarrow$ & Word Mover Distance \cite{wmd} measures dissimilarity between \emph{S,T} as the  minimum distance between their embedded words. &  \citet{metric-survey} states that WMD correlates best with human-evaluations on semantic similarity \\
 & SIM $\uparrow$ & SIM \cite{sim} uses an embedding model proposed in \cite{wieting-etal-2019-simple} to measure semantic similarity & \cite{divpara,reinforce-st} \\
\hline
\multirow{1}{*}{\makecell[l]{Style\\ Transfer (S.T.)} }& Style Accuracy $\uparrow$ & Score of 4-way and 2-way fine-tuned RoBERTa-Large \cite{roberta} model for styles in CDS and Author-Imitation datasets respectively & \cite{divpara}*,\cite{neuralcontrolledtext,adversarial-disentanglment,prototypeediting-madaan}*\\ \hline
\textbf{C.P. \& S.T.} & \textbf{Weighted Style Accuracy} $\uparrow$ & Style Accuracy weighed by their corresponding semantic similarity scores averaged across all test instances.  & \cite{divpara}*,\cite{delete-retrieve}*\\ 
\hline 
\end{tabular}
}
\caption{{\bf Evaluation Metrics:} Source and Target sentences denoted as \emph{S},\emph{T} respectively. We use all the dimensions except AMR similarity to measure the performance of TST. Note that Weighted Style Accuracy is the only metric that encompasses two crucial dimension : C.P. and S.T. in one metric. AMR similarity is used to select best performing TSTAR-Encoder. * represents the works that use slight variations of the mentioned metrics. Note that across all metrics we compute an average scores across all test instances.}
\label{tab:metrics}
\end{table*}
\begin{figure}[h]
    \centering
    \includegraphics[width=\columnwidth]{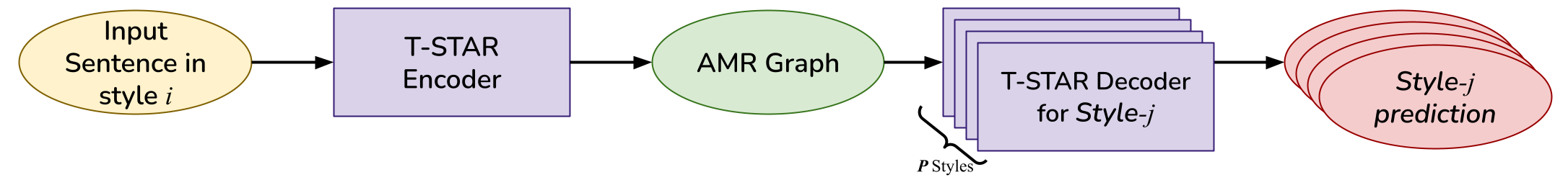}
    \caption{An overview of T-STAR model architecture. It consists of two modules: T-STAR Encoder, that transforms a given sentence $s_i$ in style $i$ to its AMR representation $A_i$. To convert the sentence to style $j$, $A_i$ is passed to T-STAR Decoder specific to style $j$.}
    \label{fig:model_figure}
\end{figure}

\section{Proposed Solution}\label{sec:solution}
Our proposed model T-STAR consists of two modules (refer Figure \ref{fig:model_figure}). First, T-STAR Encoder  generates an AMR  given source sentence in style $i$. Second, T-STAR Decoder generates a sentence in style $j$ with similar meaning as preserved in the generated intermediate AMR. 
We take T5-Base \cite{t5-model} pre-trained model as our basic seq2seq architecture for both the modules. In order to use AMR as a sequence in T5, we borrow the choice of  DFS Traversal from \cite{spring-model}, that thoroughly study the effect of various traversals on AMR parsing.


\begin{figure}[h]
    \centering
    \includegraphics[width=\columnwidth]{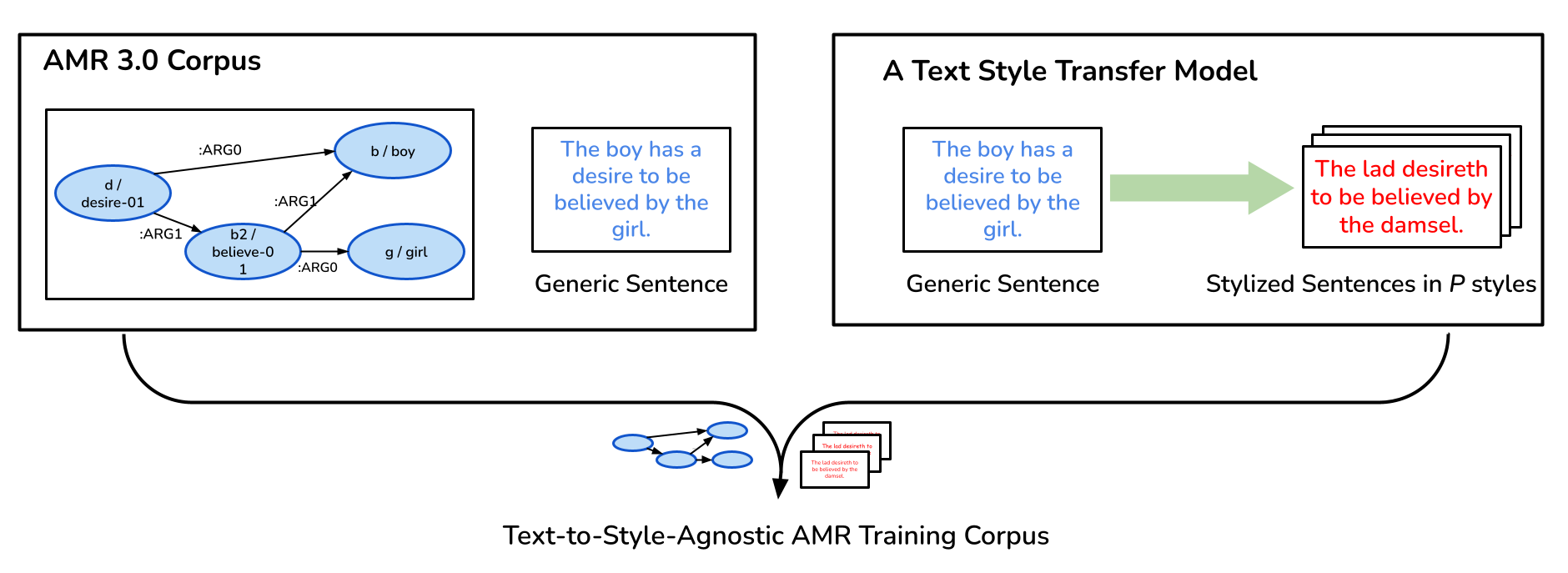}
    \caption{Generic sentences from AMR 3.0 corpus are stylized using a TST model. The corresponding AMR and stylized sentences are mapped together as silver training dataset to finetune T-STAR Encoder.}
    \label{fig:tstar_encoder}
\end{figure}

\subsection{\textbf{T-STAR Encoder}}\label{sec:tstar_encoder}
We train our simplest encoder, called the vanilla T-STAR Encoder, by fine-tuning T5 base with the  open source AMR 3.0 dataset \cite{amr3-data}. The AMR 3.0 dataset consists of roughly 59,000 generic English sentences (denoted as $s_i$ for the $i^{th}$ sentence) and their corresponding AMRs  (denoted as $A_i$).  In a qualitative analysis, we observe that the vanilla T-STAR encoder under performs in two significant ways as illustrated in Table~\ref{tab:amr-differences}.
{\bf First,} style bearing words (such as \textit{``shew''}) become concepts in the AMRs (Sentence-1 in Table~\ref{tab:amr-differences}) as opposed to being canonicalized to their respective propbank role annotations. {\bf Second,} the meaning of the stylized sentences get incorrectly encoded in the AMRs, as shown in the second example in Table \ref{tab:amr-differences}.  To overcome this, we propose a style-agnostic fine-tuning strategy as follows.

\noindent \textbf{\textit{Style-agnostic Fine Tuning}}: We hypothesize that vanilla Text to AMR encoder is unable to effectively transform stylized sentences to AMRs because it has only been trained on generic English sentences.
Therefore, we propose a data-augmentation strategy (refer to Figure \ref{fig:tstar_encoder}), where we use an off the shelf style transfer model, \textit{e.g.,} STRAP \cite{divpara}, to stylize a generic English sentence $s_i$ in style $p$ ($\hat{s}^p_i$).  While converting $s_i$ to $\hat{s}^p_i$,  we alter the style of original sentence, keeping the meaning intact. For a high quality synthetic dataset, we filter out samples with low semantic similarity between $s_i$ and $\hat{s}^p_i$. We provide a detailed empirical analysis on this filtering strategy in Appendix~\ref{encoderablation}. Since the meaning is preserved, we can now map $\hat{s}^p_i$ to the same AMR $A_i$. We then fine-tune our T-STAR Encoder on $\bar{\mathbf{S}} = \mathbf{S} \cup \mathbf{\hat{S}}$ , where $\mathbf{\hat{S}} = \{\hat{s}^p_i, A_i\}_N \forall {p \in P}$ and $P$ is the total number of styles in the dataset. 



\begin{algorithm}[t]
\small
\DontPrintSemicolon
  
  \KwInput{parallel corpora ($\mathbf{S}$), mono-lingual corpus ($\mathbf{R}^p, \forall p \in P$)}
  
  $\mathbf{\hat{S}} = \{\hat{s}^p_i, A_i\}_N \forall {p \in P}$ using an existing \textbf{TST} model, \textit{e.g.,} STRAP
  
  $X := \mathbf{S} \cup \mathbf{\hat{S}}$ 
  
  \While{Convergence}{
    \textbf{fine-tune} {TSTAR-\textit{Encoder}} $f_{amr}(.)$ on $X$
    
    $\mathbf{\hat{S}} := \{\}$
    
    \For{$p\in P$}{
    \textit{Use} $f_{amr}(.)$ to create $\mathbf{\hat{R}}^p := \{r^p_i, \tilde{A}^p_i\}_{M} $
    
    \textbf{fine-tune} TSTAR-\textit{Decoder} $f_p(.)$ for style $p$
    
    Use TSTAR-\textit{Decoder} to get $\mathbf{\hat{S}^p}=\{\hat{s}^p_i, A_i\}_N$
    
    $\mathbf{\hat{S}} = \mathbf{\hat{S}} \cup \mathbf{\hat{S}^p} $
    
    }
    $X := \mathbf{S} \cup \mathbf{\hat{S}}$
    }
\caption{Iterative T-STAR}
\label{iterative-tstar}
\end{algorithm}  
\subsection{T-STAR Decoders}

Due to the unavailability of parallel style corpora, we are provided $P$ mono-style corpora $\mathbf{R}^p = \{r^p_i\}_{M_p}$  written in style $p$, where $P$ refers to the total number of styles and $r^p_i$ refers to the $i^{th}$ sentence of the $p$ style dataset which has $M_p$ sentences.  Training a style specific decoder ($f_p(.)$) to generate a sentence in style $p$ from an AMR consists of two steps. 
{\bf First,} we use our fine-tuned T-STAR Encoder (Section~\ref{sec:tstar_encoder})  to generate AMRs $\hat{A}^p_i$ for every sentence $\mathbf{r}^p_i$ for every style corpora. {\bf Second}, we fine tune a T5 base model to recover the original sentence  $\mathbf{r}^p_i$ given the AMR $\hat{A}^p_i$, obtaining style-specific decoders. In other words, we fine tune using $M_p$ pairs of ( $\mathbf{r}^p_i$, $\hat{A}^p_i$) constructed from the first step. Note that, we experimented with a data augmentation technique for the decoders as well, however it did not lead to an improvement in the style transfer performance (refer to Appendix \ref{app:wiki}).


Once style specific decoders have been trained for every style in $P$, we can use the T-STAR Encoder in tandem with the T-STAR Decoders to convert between arbitrary style combinations as in ~\citet{divpara}.

\subsection{Iterative \textbf{T-STAR}} 
The performance of our modules, T-STAR Encoder and T-STAR Decoders  depends on the quality of the synthetic datasets ($\mathbf{\hat{S}}$, $\mathbf{\hat{R}^p}$) generated by their  complementary modules. We adopt the iterative back-translation technique used in unsupervised machine translation \cite{hoang-etal-2018-iterative}. Iteration proceeds in rounds of training an encoder and decoder from mono-style data. In every round, we aim to iteratively  improve the quality of the encoder and decoder modules by generating increasingly better synthetic data from the previous round. We briefly describe this process in Algorithm~\ref{iterative-tstar}.




\section{Experimental Setup}\label{sec:experimental_setup}
In this section we briefly describe the various T-STAR variations that are analyzed in the subsequent sections, baselines, and the implementation details. The models are validated against the metrics summarized in Table \ref{tab:metrics}.

\subsection{T-STAR variations}
\textbf{Vanilla} \textbf{T-STAR}: The T-STAR Encoder used in this version, is only trained on AMR 3.0 dataset, and not finetuned for stylized sentences.

\noindent\textbf{T-STAR}: We train the encoder and decoders using Algorithm \ref{iterative-tstar} for only one iteration.
    
\noindent \textbf{Iterative T-STAR}: We follow two iterations of Algorithm \ref{iterative-tstar} to obtain better quality synthetic dataset.

\subsection{Baselines}\label{sec:baselines}
\noindent{\textbf{UNMT}}: \cite{multiple-attribute} models style transfer as unsupervised machine translation task. 
\noindent{\textbf{DLSM}} \cite{DLSM} is a deep generative model that unifies back-translation and adversarial loss.
\noindent{\textbf{RLPrompt}} \cite{deng2022rlprompt} uses a discrete prompt optimization approach using reinforcement learning. It is adopted in a zero-shot setting where we use Distil-BERT \cite{sanh2019distilbert} and run the optimization for 1k steps.

\noindent{\textbf{STRAP}} \cite{divpara} first normalizes the style information by paraphrasing the source text into generic English sentence, which is then passed through a style-specific GPT-2 based model to generate the styled output.

\subsection{Datasets}\label{sec:datasets}
We evaluate performance of T-STAR on two English datasets that capture Linguistic Styles~\cite{styletransfer-survey}.
First, Shakespeare Author Imitation Dataset \cite{xu-etal-2012-paraphrasing} consists of $18$K pairs of sentences written in two styles. Original Shakespeare's plays have been written in Early Modern English, a significantly different style. 
Second, Corpus of Diverse Styles (CDS)~\cite{divpara}. This dataset consists of non-parallel sentences written in 11 different styles. We will present our results on a subset of four styles : Bible, Poetry, Shakespeare and Switchboard which consists of $34.8$K, $29.8$K, $27.5$K, $148.8$K instances respectively (CDS uses MIT License).

\subsection{Implementation Details}
We use pre-trained T5-Base model architecture for both the  encoder and decoder. Following iterative back translation literature \cite{kumari2021domain}, we run Iterative T-STAR for two iterations. The AMRs are preprocessed and postprocessed based on the method mentioned in Appendix  \ref{app:implementationdetails}. The best modules are selected based on the performance on validation set. Finer details about the model architecture and hyperparameters can be found in Appendix \ref{app:implementationdetails}.

 



\begin{table}[t]
\resizebox{1\columnwidth}{!}{%
\begin{tabular}{|l|l|l|l|c|}
\hline
\multirow{2}{*}{\textbf{Style}}       & \multirow{2}{*}{\textbf{Model}} & \multicolumn{2}{c|}{\textbf{Content Preservation}}  & \multirow{2}{*}{\textbf{S. R.} $\uparrow$} \\
                             &                        &  \textit{WMD} $\downarrow$ & \textit{SIM} $\uparrow$          &                                  \\ \hline\hline
\multirow{2}{*}{\textit{Bible}} 
                             & STRAP                      & 0.200 & 0.715  & 0.979                            \\
                             & T-STAR             & \textbf{0.170}  & \textbf{0.792}  & 0.979                            \\ \hline
\multirow{2}{*}{\textit{Poetry}}
                             & STRAP                      & 0.290 & 0.664  & 0.969                            \\
                             & T-STAR            & \textbf{0.215}  & \textbf{0.760 } & 0.965                            \\
                             \hline
\multirow{2}{*}{\textit{Shakespeare}} 
                             & STRAP                     & 0.328  & 0.610  & 0.971                            \\
                             & T-STAR             & \textbf{0.222}  & \textbf{0.754}  & 0.972                            \\
                             \hline
\multirow{2}{*}{\textit{Switchboard}} 
                             & STRAP                     & 0.222  & 0.751  & 0.999                            \\
                             & T-STAR             & \textbf{0.163}  & \textbf{0.848} & 1.0                             \\ \hline\hline
\end{tabular}
}
\caption{Reconstructing the original sentences using the intermediate semantic representation outperforms the baseline with a significant margin with respect to content preservation. \textbf{T-STAR} is on par with style retention (S.R.) with the baseline model.}
\label{tab:cpnumbers}
\end{table}

\section{Robustness of AMRs for Text Style Transfer}
\label{sec:amr-robustness}
 An ideal style agnostic intermediate representation for TST should a) encode complete semantic information and b) minimal style information. We quantitatively measure AMR's efficacy in these two dimensions. We compare T-STAR with STRAP~\cite{divpara} for it uses a human readable intermediate representation as well. 
 
\subsection{Semantic Containment of AMR}
If semantics of input sentence is completely preserved in the  intermediate representation, we should be able to reconstruct the input sentence from it. To evaluate the robustness of  AMRs across all styles for content preservation, we first generate intermediate AMR given the sentence in style $p$ using our encoder and then reconstruct the same sentence using our decoder $f_p(.)$ in style $p$. We study how close   the generated sentence (from STRAP, T-STAR) is with respect to the original sentence.  
We can infer from Table \ref{tab:cpnumbers} that AMRs as an intermediate representation performs significantly better on content preservation as compared to STRAP across all the styles. Specifically, it gives an average of $0.10$ and $0.06$ absolute improvement on SIM and WMD scores across the four styles respectively. Also, we get comparable performance on retaining the original style. We present an ablation study on AMR parser in Appendix \ref{encoderablation}, and propose a new unsupervised Text-AMR evaluation metric to measure the content preservation of AMR parsing along with its results on the CDS dataset in Appendix \ref{app:wmd}.

\begin{table}[t]
\centering
\resizebox{\columnwidth}{!}{%
\begin{tabular}{|l|l|l|l|l|}
\hline
\multirow{2}{*}{\textbf{Style}} & \multirow{2}{*}{\textbf{Original}} & \multirow{2}{*}{\textbf{Paraphrase}} & 
\multicolumn{2}{c|}{\textbf{AMR}} \\ \cline{4-5}
& & & Vanilla-T-STAR & T-STAR \\
 \hline
\textit{Bible}    &          97.72   &                 79.97  &     74.64  &     \textbf{71.90}       \\ \hline
\textit{Switchboard }   &     99.47        &    89.79                & 87.74       & \textbf{64.66}    \\ \hline
\textit{Poetry}    &    83.17          &    81.04                &    \textbf{49.05}      &      61.61  \\ \hline
\textit{Shakespeare}    &   \textbf{70.07}           &  78.07                  &   88.8   &   80.73   \\ \hline \hline
\textbf{Average}   &   87.61           &  82.22                 &   75.06   &   \textbf{69.73}   \\ \hline
\end{tabular}}
\caption{Accuracy of style classifiers trained on original, paraphrase and AMR inputs. AMR has least performance denoting it encodes least style information.  
}
\label{tab:styleaccuracy}
\end{table}

\subsection{Style Agnosticity of AMR} \label{sec:style-agnostic}
A style classifier $C$ assigns a style class label to an input sentence $S$. If $S$   does not encode style information, it should result in poor classifier performance. We use this observation to design an experiment to evaluate the style-agnosticity of AMRs as follows. We train three versions of 4-way style classifier using original sentences, paraphrased sentences (as used in STRAP) and AMRs as the input sequences to the classifier. Two models are used to generate AMRs: Vanilla T-STAR Encoder and T-STAR Encoder. For all versions of the style classifier, we mask content bearing words like  
  entities, numbers, common nouns and  AMR tags   in the input sequences. 
  
  The accuracy  of the 4 classifiers is shown in Table~\ref{tab:styleaccuracy}.   First, even Vanilla T-STAR Encoder has lower classifier accuracy for three out of four styles in comparison to original and paraphrases sentences. 
  Second, with the T-STAR Encoder, we observe further reduction in classifier's performance, and obtain an average absolute drop of 15.19\% and 7.1\% as compared to paraphrase and Vanilla T-STAR Encoder respectively.

 We illustrate some examples in Table \ref{tab:amr-differences}, where the T-STAR Encoder generates better AMRs as compared to the Vanilla T-STAR Encoder. In the first example,the T-STAR Encoder, unlike the vanilla T-STAR Encoder, is able to map style-specific word ``\textit{shew}'' to the valid concept ``\textit{show}''  and also associate ``\textit{existence}'' to it. In the second example, the T-STAR Encoder does not split the AMR into two sentences while parsing. Additionally, it is able to make the association between ``\textit{you}'' and ``\textit{youth}''. 
  Through the above quantitative and qualitative analysis we demonstrate that the T-STAR Encoder generates AMRs which are robust in preserving meaning of the source sentence while predominantly losing style information.

\section{Performance on Style Transfer Tasks}\label{sec:tst_perf}
 In this section, we compare our model performance T-STAR and Iterative-T-STAR to the baselines across two datasets: Shakespeare Imitation Dataset and CDS dataset.
 
\begin{table}
\resizebox{\columnwidth}{!}{%
\begin{tabular}{|l|c|c|c|c|c|}
\hline
\multirow{2}{*}{\textbf{Model}} & \multicolumn{2}{l|}{\textbf{Content Preserv.}} & \textbf{Lex. Div.}                                                                                                                   & \multicolumn{1}{l|}{\multirow{2}{*}{\textbf{\begin{tabular}[c]{@{}l@{}} S.T.\end{tabular}}}} & \multicolumn{1}{l|}{\multirow{2}{*}{\textbf{WSAcc.}}} \\ \cline{2-4}

                                & \multicolumn{1}{l|}{\textit{SIM} $\uparrow$} & \multicolumn{1}{l|}{\textit{WMD} $\downarrow$} & \multicolumn{1}{l|}{\textit{Self-BLEU} $\downarrow$} & \multicolumn{1}{l|}{}                                                                                   & \multicolumn{1}{l|}{}                                 \\ \hline
                                \multicolumn{6}{|c|}{\textbf{\textit{Original to Modern Style}}}\\\hline
UNMT                   & \multicolumn{1}{c|}{0.461}        & \multicolumn{1}{c|}{0.318}        & \multicolumn{1}{c|}{0.118}                      & 0.586                                                                                                   & 0.256                                                 \\ 
DLSM                   & \multicolumn{1}{c|}{0.447}        & \multicolumn{1}{c|}{0.369}        & \multicolumn{1}{c|}{\underline{0.079}}                     & 0.192                                                                                                   & 0.115                                                 \\
RLPrompt                   & \multicolumn{1}{c|}{0.508}        & \multicolumn{1}{c|}{0.387}        & \multicolumn{1}{c|}{\underline{0.164}}                     & 0.354                                                                                                   & 0.292                                                 \\
{STRAP}                  & \multicolumn{1}{c|}{0.647}        & \multicolumn{1}{c|}{0.337}        & \multicolumn{1}{c|}{0.118}          & \underline{0.886 }                                                                                                  & 0.552                                                 \\ \hline
Vanilla-TSTAR             & \multicolumn{1}{c|}{\underline{0.848}}        & \multicolumn{1}{c|}{\underline{0.182}}        & \multicolumn{1}{c|}{0.269}     & 0.601                                                                                                   & 0.497                                                 \\ 
TSTAR & 0.754 & 0.257 & 0.175  & 0.754 & 0.554 \\
Iterative-TSTAR             & \multicolumn{1}{c|}{0.799}        & \multicolumn{1}{c|}{0.227}        & \multicolumn{1}{c|}{0.209}    & 0.715   & \textbf{0.556}  \\ \hline
\multicolumn{6}{|c|}{\textbf{\textit{Modern to Original Style}}}\\\hline
UNMT                   & \multicolumn{1}{c|}{0.373}        & \multicolumn{1}{c|}{0.375}        & \multicolumn{1}{c|}{\underline{0.057}}                         & 0.414                                                                                                   & 0.158                                                 \\ 
DLSM                   & \multicolumn{1}{c|}{0.421}        & \multicolumn{1}{c|}{0.373}        & \multicolumn{1}{c|}{0.086}                              & 0.391                                                                                                   & 0.174                                                 \\
RLPrompt                   & \multicolumn{1}{c|}{0.550}        & \multicolumn{1}{c|}{0.348}        & \multicolumn{1}{c|}{\underline{0.261}}                     & 0.547                                                                                                   & 0.203                                                 \\
{STRAP}                  & \multicolumn{1}{c|}{0.656}        & \multicolumn{1}{c|}{0.332}        & \multicolumn{1}{c|}{0.139}                          & \underline{0.681}                                                                                                  & 0.433                                                 \\ \hline

Vanilla TSTAR & \underline{0.897} &\underline{0.14}& 0.379 & 0.47 & 0.418 \\
TSTAR             & \multicolumn{1}{c|}{0.842}        & \multicolumn{1}{c|}{0.185}        & \multicolumn{1}{c|}{0.324}       & 0.490                                                                                                   & 0.402                                                 \\ 
Iterative-TSTAR             & \multicolumn{1}{c|}{0.853}        & \multicolumn{1}{c|}{0.181}        & \multicolumn{1}{c|}{0.329}  & 0.540 & \textbf{0.446}  \\ \hline
\end{tabular}}
\caption{T-STAR models comparison with baseline models on Shakespeare Author Imitation Dataset for both the directions. Iterative T-STAR outperforms all the baselines on Weighted Style Accuracy.}
\label{authorimitation}
\end{table}


\begin{table}[t]
\resizebox{\columnwidth}{!}{%
\begin{tabular}{|l|l|l|l|l|l|l|l|}
\hline
\multirow{2}{*}{\textbf{Direction}}                             & \multirow{2}{*}{Model} & \textbf{L.D.} & \multicolumn{2}{c|}{\textbf{Cont. Preserv.}} & \multirow{2}{*}{\textbf{S.R.} $\downarrow$} & \multirow{2}{*}{\textbf{S.T.} $\uparrow$} & \multirow{2}{*}{\textbf{WSAcc} $\uparrow$} \\
                                                       &                        & \textit{S-BLEU $\downarrow$}    & \textit{WMD $\downarrow$}            & \textit{SIM $\uparrow$}           &                                  &                                 &                                          \\
\hline
\multirow{4}{*}{\textit{poetry $\rightarrow$ bible}}            & STRAP                  & \underline{0.067}     & 0.314 & 0.571 & 0.335                            & 0.548                           & 0.289                                    \\
                                                       & Van-TSTAR                   & 0.170  & \underline{0.183} & \underline{0.812} & 0.430                             & 0.289                           & 0.226                                    \\ \cline{2-8}
                                                       & TSTAR & 0.084   & 0.268   & 0.670 & \underline{0.202}                           & \underline{0.618}                           & \textbf{0.391}                           \\
                                                      & Itr-TSTAR & 0.106 & 0.241 & 0.721 & 0.231 & 0.566 & 0.383 \\ \hline\hline
\multirow{4}{*}{\textit{shak. $\rightarrow$ bible}}       & STRAP            &      \underline{0.073}   & 0.343 & 0.535 & 0.291                            & 0.677                           & 0.346                                    \\
                                                       & Van-TSTAR                   & 0.212  & \underline{0.194}  & \underline{0.817} & 0.540                            & 0.422                           & 0.337                                    \\ \cline{2-8}
                                                       & TSTAR & 0.123  & 0.286  & 0.656 & 0.277                            & \underline{0.705}                           & 0.437                                    \\
                                                       & Itr-TSTAR & 0.149 & 0.260 & 0.712 & \underline{0.271} & 0.701 & \textbf{0.477} \\ \hline\hline   
                                                       
\multirow{4}{*}{\textit{switch. $\rightarrow$ bible}}       & STRAP                  & \underline{0.042}   & 0.323 & 0.476 & 0.062                            & 0.670                           & 0.307                                    \\
                                                       & Van-TSTAR                   & 0.128  & \underline{0.196} & \underline{0.729} & 0.085                            & 0.456                           & 0.330                                    \\\cline{2-8}
                                                       & TSTAR & 0.758   & 0.260 & 0.605 & 0.007                            & 0.725                           & 0.419                                    \\
                                                       & Itr-TSTAR & 0.097 & 0.241 & 0.637 & \underline{0.006} & \underline{0.745} & \textbf{0.456} \\ \hline\hline                                                                   
\multirow{4}{*}{\textit{poetry $\rightarrow$ shak.}}      & STRAP                  & \underline{0.067}   & 0.327 & 0.571 & \underline{0.159 }                           & \underline{0.810}                            & 0.450                                    \\
                                                       & Van-TSTAR                   & 0.207  & \underline{0.166}  & \underline{0.821} & 0.398                            & 0.576                           & 0.460                                    \\\cline{2-8}
                                                       & TSTAR & 0.131  & 0.244  & 0.717 & 0.241                            & 0.733                           & 0.509                           \\
                                                       & Itr-TSTAR & 0.164 & 0.207 & 0.768 & 0.278 & 0.704 & \textbf{0.522} \\ \hline\hline
\multirow{4}{*}{\textit{switch. $\rightarrow$ shak.}} & STRAP                  & \underline{0.045}   & 0.328 & 0.489 & 0.012                            & 0.956                           & 0.461                                    \\
                                                       & Van-TSTAR                   & 0.152  & \underline{0.187}  & \underline{0.740} & 0.034                            & 0.948                           & \textbf{0.696}                                    \\ \cline{2-8}
                                                       & TSTAR & 0.089   & 0.250 & 0.651 & \underline{0.009}                            & 0.971                           & 0.628 \\
                                                       & Itr-TSTAR & 0.122 & 0.22 & 0.696 & 0.010 & \underline{0.973} & 0.675 \\
\hline  \hline                                         
\multirow{4}{*}{\textit{bible $\rightarrow$ shak.}}       & STRAP                  & 0.110  & 0.242 & 0.634 & \underline{0.231}                            & \underline{0.764}                           & 0.465                                    \\
                                                       & Van-TSTAR                   & 0.214  & \underline{0.151} & \underline{0.842} & 0.377                            & 0.613                           & \textbf{0.508}                                    \\\cline{2-8}
                                                       & TSTAR & \underline{0.100}  & 0.227   & 0.713 & 0.309                            & 0.676                           & 0.472                                    \\
                                                       & Itr-TSTAR & 0.126 & 0.210& 0.743 & 0.299 & 0.683 & 0.499 \\ \hline\hline

\multirow{4}{*}{\textit{bible $\rightarrow$ switch.}}       & STRAP                  & 0.088    & 0.238 & 0.625 & \underline{0.080}                            & \underline{0.918 }                          & 0.565                                    \\
                                                       & Van-TSTAR                   & 0.171  & \underline{0.162} & \underline{0.810} & 0.186                            & 0.813                           & 0.649                                    \\\cline{2-8}
                                                       & TSTAR & \underline{0.084}   & 0.227 & 0.713& 0.082                            & 0.916                           & 0.646                                    \\
                                                       & Itr-TSTAR & 0.101 & 0.213 & 0.73 & 0.083 & 0.911 & \textbf{0.658} \\ \hline \hline
\multirow{3}{*}{\textit{shak. $\rightarrow$ switch.}} & STRAP                  & \underline{0.039}   & 0.344  & 0.534 & \underline{0.000}                            & \underline{0.998}                           & 0.533                                    \\
                                                       & Van-TSTAR                   & 0.138  & \underline{0.212} & \underline{0.767} & 0.002                            & 0.988                           & 0.755                                    \\\cline{2-8}
                                                       & TSTAR & 0.089   & 0.283 & 0.661 & \underline{0.000}                            & \underline{0.998}                           & 0.660                                    \\
                                                       & Itr-TSTAR & 0.110 & 0.249 & 0.719 & 0.002 & 0.991 & \textbf{0.712} \\\hline          
\multirow{4}{*}{\textit{poetry $\rightarrow$ switch.}}      & STRAP                  &\underline{0.054}   & 0.312 & 0.616 & \underline{0.007}                            & \underline{0.993}                          & 0.610                                    \\
                                                       & Van-TSTAR                   & 0.164  & \underline{0.184}  & \underline{0.825} & 0.028                            & 0.972                           & \textbf{0.801}                           \\\cline{2-8}
                                                       & TSTAR & 0.113  & 0.250  & 0.740 & 0.015                            & 0.985                           & 0.728                                    \\
                                                       & Itr-TSTAR & 0.141 & 0.213 & 0.791 & 0.019 & 0.981 & 0.774 \\ \hline\hline
                                                
\multirow{4}{*}{\textit{bible $\rightarrow$ poetry}}            & STRAP                  & 0.060   & 0.275  & 0.633  & 0.346                            & \underline{0.621}                           & 0.374                                    \\
                                                       & Van-TSTAR                   & 0.094   & \underline{0.195} & \underline{0.777}  & 0.533                            & 0.417                           & 0.315                                    \\\cline{2-8}
                                                       & TSTAR & \underline{0.054}   & 0.253 & 0.684 & \underline{0.323}                            & 0.597                           & \textbf{0.394}                           \\
                                                       & Itr-TSTAR & 0.064 & 0.237 & 0.700 & 0.348 & 0.559 & 0.375 \\ \hline\hline

\multirow{4}{*}{\textit{shak. $\rightarrow$ poetry}}      & STRAP                  & \underline{0.071}   & 0.353  & 0.550 & \underline{0.160}                            & \underline{0.818}                           & 0.449                                    \\
                                                       & Van-TSTAR                   & 0.155  & \underline{0.219}  & \underline{0.753} & 0.228                            & 0.748                           &\textbf{0.551}                                   \\ \cline{2-8}
                                                       & TSTAR & 0.108  & 0.286  & 0.641 & 0.166                            & 0.815                           & 0.512                                    \\
                                                       & Itr-TSTAR & 0.128 & 0.258 & 0.693 & 0.190 & 0.794 & 0.538 \\ \hline\hline

\multirow{4}{*}{\textit{switch. $\rightarrow$ poetry}}      & STRAP                  & \underline{0.037}   & 0.331  & 0.524 & \underline{0.094}                            & \underline{0.844}                           & \textbf{0.432}                                    \\
                                                       & Van-TSTAR                   & 0.101  & \underline{0.215} & \underline{0.715} & 0.261                            & 0.486                           & 0.334                                    \\\cline{2-8}
                                                       & TSTAR & 0.064   & 0.268 & 0.631 & 0.106                            & 0.701                           & 0.427                                    \\
                                                       & Itr-TSTAR & 0.084 & 0.239 & 0.673 & 0.158 & 0.603 & 0.388 \\ \hline\hline\hline

\multirow{4}{*}{\textbf{All Styles Avg.}}      & STRAP                  &  \underline{0.063} &  0.311 & 0.563 & 0.148 & \underline{0.801}    & 0.440                                 \\
                                                       & Van-TSTAR                   & 0.159  & \underline{0.189} & \underline{0.784} & 0.259 & 0.644 & 0.497 \\\cline{2-8}
                                                       & TSTAR & 0.093  & 0.256 & 0.674 & \underline{0.145} & 0.787 & 0.519 \\
                                                       & Itr-TSTAR & 0.116 & 0.232 & 0.715 & 0.158 & 0.768 & \textbf{0.538} \\ \hline

\end{tabular}}
\caption{Performance comparison of T-STAR models with STRAP on 12 different directions, across four styles. T-STAR and Iterative T-STAR beats STRAP for 11 directions out of 12. S.R. - Style Retention, S.T. - Style Transfer, WSAcc - Weighted Style Accuracy.}
\label{st-numbers}
\end{table}

\subsection{Performance Analysis on Shakespeare Imitation Dataset} \label{sec:shakes-results}
In Table~\ref{authorimitation}, we present the results of T-STAR and Iterative-T-STAR in comparison with the baselines for the Shakespeare Author Imitation dataset for both the directions :original to modern style and vice versa. The Weighted Style Accuracy is the primary metric as it is shown to effectively combine both style accuracy and content preservation~\cite{divpara}.

We observe that our model, T-STAR slightly performs better than STRAP model for original to modern style, but has  lower performance for modern to original style. However Iterative T-STAR, outperforms all the baselines for both the directions on Weighted Style Accuracy. We observe that STRAP has very high style accuracy that it achieves by compromising on Content Preservation. Through human evaluation (Section~\ref{sec:human_eval}) we see STRAP employs significantly higher hallucinations to achieve style transfer. Vanilla T-STAR on the other hand achieves high content preservation via significant copying from source  as seen by the substantially high self-BLEU score. Iterative T-STAR finds the middle ground of achieving style transfer while not compromising on content preservation. We also found that the length of generated sentence is similar to the input sentence, with on average one word difference. In the subsequent sections, we compare T-STAR to only the best performing baseline, STRAP. 
%




\subsection{Performance Analysis on CDS Dataset} \label{sec:cds-results}
In Table~\ref{st-numbers}, we compare the performance of T-STAR and Iterative T-STAR against STRAP and Vanilla T-STAR, across all 12 directions for \textit{\{Poetry, Shakespeare, Switchboard, Bible\}} styles. 
We make the following observations: \textbf{First}, both our models T-STAR and Iterative T-STAR outperform the state-of-the-art baseline, STRAP, on 11 out of 12 directions, with an average absolute improvement of 7.7\% and 9.7\% respectively. \textbf{Second}, Vanilla T-STAR is observed to be a stronger baseline than STRAP, as it beats STRAP on 8 out of 12 directions. \textbf{Third},  when we compare iterative T-STAR against vanilla T-STAR, we beat it in 7 out of 12 directions, where the average absolute improvement across these 7 directions is 8\%, as compare to the average absolute loss in 5 out of 12 styles is 2\%. \textbf{Fourth}, Iterative T-STAR model outperforms T-STAR model on 9 out of 12 directions, which shows that the improvement in synthetic dataset quality, is boosting the performance of the downstream task. Consistent with the findings of the previous experiment, we see Iterative T-STAR is able to find the middle ground of transferring style without compromising on content preservation.

\begin{table*}[t]
\resizebox{1\textwidth}{!}{
\begin{tabular}{|p{2.5cm}|p{3.5cm}|p{3.5cm}|p{3.5cm}|p{3.5cm}|p{3.5cm}|}
\hline
\textbf{Direction} & \textbf{Input Sentence }                                                      & \textbf{STRAP}                                     & \textbf{Vanilla T-STAR }                      & \textbf{T-STAR}                                               & \textbf{Iterative T-STAR  }                        \\\hline
\textit{Poetry}$\rightarrow$\textit{Bible} & ``Dear Lord! It has a fiendish look"-- & But they laughed, and said, Oh my God!    & Thou lookest fiendish, O LORD        & Behold, it beholdeth to be fiendish, saith the LORD. & Dear Lord, it looketh fiendish unto thee. \\\hline
\textit{Bible}$\rightarrow$\textit{Poetry} & And he said unto another, Follow me.  & And bade the other follow me;             & And another: ``O follow me!" said he, & And follow me! and I will be be ye gone,             & And thou, my love, will follow me.        \\\hline
\textit{Shake}.$\rightarrow$\textit{Switch}. & Dead art thou, dead!                & did you get a uh a uh you have a dead art & you're dead i say you are dead       & oh you'd die                                         & you're dead yeah                          \\\hline
\textit{Shake}.$\rightarrow$\textit{Switch}.& Thus with a kiss I die.             & so i'm i'm dying to get a kiss            & so i die with a kiss                 & i die with a kiss                                    & so i die with a kiss                     \\ \hline
\end{tabular}}
\caption{Example of generated stylized sentences for STRAP, Vanilla T-STAR, T-STAR and Iterative T-STAR models for the given input sentence.}
\label{tab:case-study}
\end{table*}

\subsubsection{Qualitative Analysis}
In Table \ref{tab:case-study}, we enumerate few examples with generated stylized sentences using STRAP and T-STAR variations. We can infer that, although STRAP performs well in transforming the sentence to the given style, it alters the meaning (row 1,3, and 4). On the other hand, vanilla T-STAR does not always transform the style (row 1 and 2). However, with T-STAR and Iterative T-STAR are able to transform the sentences while keeping the style intact. 

We further quantify these observations through an extensive set of human annotations as described below.

 \section{Human Evaluations}\label{sec:human_eval} 
\begin{table}[]
\centering 
\resizebox{0.8\columnwidth}{!}{
\begin{tabular}{|l|c|c|c|}
\hline
\textbf{Model }&  $>$ STRAP&  $<$ STRAP &  = STRAP \\             \hline
\textit{T-STAR} &  70.8\% &  26.4\% &  2.8\% \\\hline
\textit{Itr. T-STAR} & 77.5\% & 20.5\% & 2.0\%       \\ \hline
\end{tabular}}
\caption{Comparison of T-STAR and Iterative T-STAR models against STRAP for content preservation.}
\label{agg-cp-analysis}
\end{table}

Automatic metrics are insufficient to thoroughly understand subjective quality measures like content preservation. Therefore, we conduct an extensive case study with human evaluations. Our analysis is two folds, first we compare STRAP with our models T-STAR and Iterative T-STAR on meaning preservation.
Second, we further understand the various categories of meaning loss failures. 

For both the human evaluation tasks below, the criteria for choosing annotators were i) proficient in English with a minimum education of  Diploma. ii) The annotators have to first qualify on two simpler questions, else they are not allowed to continue on the task. Each instance is annotated by three taskers, and the final annotation is a majority vote. 


\subsection{Comparison on Meaning Preservation}\label{sec:meaning_preservation}
In order to study the faithfulness of the T-STAR models, we do a side-by-side human evaluation. In this task, a source sentence is shown with 2 stylized target sentence (one from T-STAR and another from STRAP).   We present the annotators with three options to judge content preservation with respect to source sentence:   option on left better than one on right, right better than left and both equal.
  
We extensively compare the two models across all 12 directions for   four styles.
For each direction we randomly sample 500 instances. Each instance is rated by 3 annotators leading to a total of 18,000 ratings. We   summarize our findings in Table \ref{agg-cp-analysis}. Both T-STAR ad Iterative T-STAR significantly outperform STRAP in terms of being better at content preservation (The $>$ STRAP column in Figure~\ref{agg-cp-analysis}). Further Iterative T-STAR has 7\% higher meaning preservation compared to T-STAR. In addition to the quantitative content preservation metrics discussed in Section~\ref{sec:tst_perf}, this analysis  gathers additional qualitative evidence towards AMRs as an effective intermediate representation for  content preserving style transfer.    The complete statistics per direction are available in Appendix \ref{app:erroranalysis}.

\begin{table}[t]
\resizebox{1\columnwidth}{!}{
\begin{tabular}{|l|llll|}
\hline
\multirow{2}{*}{\textbf{Model}} & \multicolumn{4}{c|}{\textbf{Type of Error}}                                                                                                                                                   \\ \cline{2-5} 
                                & \multicolumn{1}{l|}{\textit{\textbf{None}}$\uparrow$}&  \multicolumn{1}{l|}{\textit{\textbf{Hal.}}$\downarrow$}  & \multicolumn{1}{l|}{\textit{\textbf{Sem. Drift.}}$\downarrow$}  & \textit{\textbf{Incomp.}$\downarrow$}   \\ \hline
\textit{STRAP}                  & \multicolumn{1}{l|}{10.48\%}                        & \multicolumn{1}{l|}{39.38\%}                            & \multicolumn{1}{l|}{35.38\%}                             & 14.75\%                          \\ \hline
\textit{T-STAR}                 & \multicolumn{1}{l|}{19.36\%}                       & \multicolumn{1}{l|}{24.46\%}                            & \multicolumn{1}{l|}{33.71\%}                             & 22.45\%                         \\ \hline
\textit{Itr. T-STAR}       & \multicolumn{1}{l|}{24.83\%}                       & \multicolumn{1}{l|}{22.6\%}                            & \multicolumn{1}{l|}{36.98\%}                             & 15.58\%                          \\ \hline
\end{tabular}}
\caption{Aggregate Error Analysis for error types: Hallucinations, Semantic Drift and Incomplete for 6,000 samples. With Iterative T-STAR, the number of samples with no errors and less hallucinations increase significantly.}
\label{agg-error-analysis}
\end{table} 
\subsection{Error Analysis}\label{sec:error_analysis} 
In the next study, we further aim to study the nature of meaning loss errors made by style tranfer models. We categorize these errors into three categories  
i) \textit{Hallucinations}: new information not present in the source sentence is added to target ii) \textit{Semantic Drift:} the target sentence has a different meaning to source sentence iii) \textit{Incomplete:} some important content information is missed in the target. 
The taskers also have the option to select ``No Error" if the meaning is preserved in the generated target. As in the previous experiment, we collect 18,000 ratings and the results are summarized in  Table~\ref{agg-error-analysis}.

We observe that our models T-STAR and Iterative T-STAR consistently beat STRAP in the \textit{``No Error''} category. Furthermore, the amount of hallucinations significantly drops to 24.46\% and further 22.6\% with Iterative T-STAR across all styles from  39.3\% for STRAP. Reduction in hallucination can be clearly seen as a benefit of encoding critical information in the source sentence using a semantic parse representation like AMR. As a sign of improving the AMR parsing quality, we see that iterative T-STAR further reduce the Incomplete to 15.5\%. For further details refer to Appendix \ref{app:erroranalysis}.

\subsubsection{Usefulness of an interpretable intermediate representation}
With intermediate AMRs being interpretable, it is possible to broadly understand if such errors are emerging from either encoder or decoder module. To intuitvely understand the reason for high number in \textit{Incomplete} and \textit{Semantic Drift} errors, we qualitatively analyzed some instances along with the generated intermediate AMRs. We have listed down these examples in Appendix \ref{app:erroranalysis}. We observed that for \textit{Incomplete} errors, the generated AMRs were not encoding complete information, and thus this error percolated from the T-STAR encoder. For the majority of the instances, either some entities were missing, and if the clause was separated using ``:,;'', it was not parsed in the intermediate graph. \textit{Semantic Drift} errors indicate shortcomings in both the modules, for some instances the encoder is not abstracting out the meaning efficiently and for others the decoder is not able to generate sentences with the meaning encoded in the intermediate AMR.

\section{Conclusion} 
We explored the use of AMR graphs as an intermediate representation for the TST task. We see that the performance of the proposed method T-STAR surpasses  state of the art techniques in content preservation with comparable style accuracy.  Through qualitative analysis we show that obtaining very high style accuracy scores without altering meaning is indeed a challenging problem.

  \section{Limitations} \label{sec:limitations} 
 Some of the limitations for T-STAR based models are the following. \textbf{First}, although our proposed models are performing better in the joint objective of content preservation and style transfer, but they are not able to outperform vanilla T-STAR (overall best performing model for CP) and STRAP (overall best performing model for ST). This is a promising future direction, to keep boosting the performance on both the directions without comprising on the other dimension. \textbf{Second}, we are not incorporating graph structure in our models, and thus there could be some information loss while interpreting and generating the AMRs. \textbf{Third,} based on our error analysis,  although our T-STAR encoder is able to generate better AMRs for stylized sentences as compared to vanilla T-STAR model, we are generating significant incomplete AMRs that are missing out on important entities and relations to preserve meaning of source sentence. \textbf{Fourth}, similar to prior research to generate synthetic dataset, initial iteration of our model are dependant on an existing off the shelf TST model, however the quality of the generated AMRs improves significantly using the described data augmentation strategy. \textbf{Fifth}, our work is dependant on a robust AMR parsing approach, which makes it challenging to adopt our approach for other languages. However, with the recent advancements in multilingual AMR parsing, it will be feasible in upcoming future works.

\bibliographystyle{acl_natbib}
\bibliography{anthology,acl2021}

\clearpage
\appendix

\section{Extended Related Works}
\label{app:extendrw}
\subsection{TST Metrics} 
Table~\ref{tab:metrics} summarizes different metrics that have been used to measure content preservation and style transfer efficacy. 
\citet{metric-survey} presents a comprehensive analysis and categorization of several such metrics with respect to human evaluations. \citet{tikhonov2019style} also points out some flaws in traditional evaluation techniques and insists on using human written reformulations for better evaluation\footnote{Due to the difficulty of our task, we instead restrict to human evaluations instead of obtaining human gold-standard benchmarks.}.

\subsection{Text to AMR }
Recent works \cite{spring-model,cai-lam-2020-amr,zhou2020amr} for text-to-AMR task have pushed the SOTA, that makes it feasible to automatically construct AMR given a sentence. As a consequence, semantic-preserving NLG tasks, such as Neural Machine Translation \cite{amr-nmt, nmt-ucca}, Abstractive Summarization \cite{amr-summarization}, and Question Decomposition for multi-hop question answering \cite{deng2022interpretable} use AMRs as intermediate representations. However, AMRs have not been explored for style transfer tasks before our work. 

The increase in AMRs being adopted for several seq2seq tasks is due to the boost in the quality of AMR parsers. Earlier works, relied on statistical approaches \cite{peng-etal-2017-addressing,flanigan-etal-2014-discriminative, flanigan-etal-2016-cmu} to generate AMRs for a given piece of text. With the emergence of deep learning, various AMR parsers are being proposed, which can be divided into following categories: i) sequence-to-sequence based AMR-parsers \cite{xu2020improving}, ii) sequence-to-graph based AMR parsers \cite{zhang2019amr}, where the graph is incrementally built by spanning one node at a time. More recently, several works have adopted pretrained models for AMR parsers, and have observed a boost in performance. \citet{bai2022graph} uses BART model and posit the AMR parsing task as a seq2seq task, and generates a traversal of the AMR parser as the output. \citet{bai2022graph} incorporates a pretraining strategy to better encode graph information in the BART architecture. \citet{xu2020improving} uses sentence encoding generated from BERT model. In this work, we adopt the pretrained technique based AMR parser, to generate high quality AMRs for the given stylized sentences. 

Although off-the-shelf AMR parsers work well for some problems \cite{fan2020multilingual}, they often need to be modified to be useful in the downstream tasks. For instance, \citet{deng2022interpretable} proposed graph segmentation strategy to perform question decomposition on a multi-hop query. 

\citet{xia2021stacked} and \citet{du2021avoiding} illustrated that using silver data augmentation can help improve in the task of AMR parsing. In this work, we also illustrate the benefit of using silver data towards improving the style agnosticity  of AMR graphs as an intermediate representation. 

\subsection{AMR to Text}

Similar to text-to-AMR models, AMR-to-text frameworks can also be categorised into two types - i ) sequence-to-sequence generation frameworks \cite{}, ii) graph-encoder based frameworks \cite{song2018graph, wang2021better, wang2020amr}. \citet{bai2020online} propose a decoder that back-predicts projected AMR graphs to better preserve the input
meaning than standard decoders. \citet{bai2022graph} argues that PLMs are pretrained on textual data, making is sub-optimal for modeling structural knowledge, and hence propose self-supervised graph-based training objectives to improve the quality of AMR-to-text generation.

\section{Implementation Details} 
\label{app:implementationdetails}

\textbf{Offensive language}
We used the ``List of Dirty, Naughty, Obscene
or Otherwise Bad Words"\footnote{\hyperlink{https://github.com/LDNOOBW/List-of-Dirty-Naughty-Obscene-and- Otherwise-Bad-Words}{https://github.com/LDNOOBW/List-of-Dirty-Naughty-Obscene-and-Otherwise-Bad-Words}} to validate that the source and the generated target text do not contain any offensive text.

\textbf{Model Architecture:}
We use a standard \texttt{t5-base} encoder-decoder model as described in \shortcite{t5-model}.
The pre-trained \texttt{HuggingFace}\footnote{\hyperlink{https://github.com/huggingface/transformers}{https://github.com/huggingface/transformers}} T5 transformer is used for both text-to-AMR and AMR-to-text parts of the proposed architecture. The model is pre-trained on the Colossal Clean Crawled Corpus (C4) \footnote{\hyperlink{https://www.tensorflow.org/datasets/catalog/c4}{https://www.tensorflow.org/datasets/catalog/c4}} comprising of $\sim$750 GBs worth of text articles. The model comprises of 220 billion parameters, and is pre-trained for $2^19$ steps before fine-tuning. For pre-training, AdaFactor optimizer \cite{adafactor-optimizer} is used with ``inverse square root" learning rate schedule.

\textbf{AMR Graph Construction:} We use the SPRING model \cite{spring-model} to generate AMR graphs from source style text. We use \texttt{amrlib}\footnote{\hyperlink{https://github.com/bjascob/amrlib}{https://github.com/bjascob/amrlib}} package to generate the AMR graphs. This implementation uses \texttt{T5-base} \cite{t5-model} as its underlying model, as opposed to the \texttt{BART model} \cite{bart-model} in the SPRING architecture.
It is trained on \textbf{AMR 3.0} (LDC2020T02) dataset \cite{amr3-data} that consists of 59K manually created sentence-AMR pairs. The model is trained for 8 epochs using a learning rate of $10^{-4}$. The source and target sequence lengths are restricted to 100 and 512 tokens respectively.
Note that t5-based SPRING model achieves an SMATCH score of $83.5$, which guarantees the quality of obtained AMR representations $z_i$.

\textbf{AMR-Based Style Transfer:}
We use the \texttt{T5wtense} (T5 with tense) architecture from the \texttt{amrlib} package. The \texttt{T5wtense} architecture encodes part-of-speech (POS) tags to the concepts in the AMR graph, which helps the generation model to predict the tense of the output sentence since AMR graphs do not retain any tense information from their corresponding sentence. This model outperforms the standard T5-based model by 10 BLEU points on the \textbf{AMR 3.0} (LDC2020T02) dataset \cite{amr3-data}. To keep the training steps comparable for the subsets of the CDS dataset \cite{divpara}, we train this \texttt{t5-base} model for $20$ epochs for the Bible, Romantic Poetry, and Shakespeare datasets, and $5$ epochs for the Switchboard dataset. The model was trained for $20$ epochs for the Shakespeare Author Imitation Dataset \cite{xu-etal-2012-paraphrasing} as well. We used a learning rate of $10^{-4}$ for both datasets, and restricted source and target sequence lengths to 512 and 90 throughout, respectively. Everything else was kept same as the \texttt{amrlib} implementation to keep the results consistent.

\textbf{STRAP baseline:}
We train the model keeping the same hyperparameter configuration as reported in \citet{divpara}. We train each style-specific decoder for 3 epochs with learning rate of $5 \times 10^{-5}$ with Adam optimizer \cite{adam-optimizer}. During inference we set the \textit{p}-value for nucleus sampling \cite{nucleus-sampling} to 0.7 to have an appropriate balance between content preservation and style accuracy scores.

\textbf{Style classifiers:}
Similar to \citet{divpara}, we fine-tune a \texttt{RoBERTa-large} model \cite{roberta} using the official implementation in the \texttt{fairseq} package\footnote{\hyperlink{https://github.com/pytorch/fairseq}{https://github.com/pytorch/fairseq}} to train the style classifiers mentioned in Table \ref{tab:cpnumbers}. For all classifier variants, learning rate of $10^{-5}$ and a mini-batch size of $32$ was used. The models were trained for 10 epochs using Adam optimizer \cite{adam-optimizer}. We also masked out named entities, nouns and numbers from the input before training the classifier. We used \texttt{spacy} package\footnote{\hyperlink{https://spacy.io/}{https://spacy.io/}} to obtain the named entities and POS tags. To obtain the named-entities and POS for AMR graphs, information extracted from original sentences was used.

\textbf{Train-Validation-Test splits:}
The data splits were kept the same as the baseline model, STRAP \cite{divpara} and can be found in Table \ref{tab:data-split}. 

\begin{table}[htb]
\centering
\resizebox{0.8\columnwidth}{!}{%
\begin{tabular}{|l|l|l|l|l|}
\hline
\multicolumn{2}{|l|}{\textbf{Dataset}} & 
\textbf{Train split} & \textbf{Validation split} & \textbf{Test Split} \\ \hline
\multirow{4}{*}{CDS} & Bible & 31,404 & 1,714 & 1,714 \\
& Switchboard & 145,823 & 1,487 & 1,488 \\
& Poetry & 26,880 & 1,464 & 1,470 \\
& Shakespeare & 24,852 & 1,313 & 1,293 \\
\hline
\multirow{2}{*}{SAID} & Original & 18,395 & 1,128 & 1,462 \\
& Modern & 18,395 & 1,218 & 1,462 \\
\hline
\end{tabular}}
\caption{Size of train, validation and test sets for CDS dataset \cite{divpara} and Shakespeare Author Imitation Dataset (SAID) \cite{xu-etal-2012-paraphrasing}.}
\label{tab:data-split}
\end{table}

\textbf{Computational time and device setup:}
All experiments were done on a 16 GB NVIDIA V100 GPU system with 120 GB n1-standard-32 Intel Broadwell CPU. It took $\sim$16 hrs to train the AMR-to-Text models for Shakespeare Author Imitation dataset \cite{xu-etal-2012-paraphrasing} and $\sim$25 hrs for the CDS dataset. 

\textbf{Evaluation Metrics:}
We used the \texttt{gensim}\footnote{\hyperlink{https://radimrehurek.com/gensim/}{https://radimrehurek.com/gensim/}} package to compute the Word Mover Distance (WMD) \cite{wmd}, \texttt{nltk}\footnote{\hyperlink{https://www.nltk.org/}{https://www.nltk.org/}} package to compute the BLEU scores \cite{bleu}, \texttt{smatch} package\footnote{\hyperlink{https://github.com/snowblink14/smatch}{https://github.com/snowblink14/smatch}} to compute the SMATCH score \cite{smatch} and implementation by Krishna et al. \shortcite{divpara}\footnote{\hyperlink{https://github.com/martiansideofthemoon/style-transfer-paraphrase}{https://github.com/martiansideofthemoon/style-transfer-paraphrase}} for the cosine similarity using SIM embeddings \cite{sim}.

\textbf{License of the packages used:}
The following packages use the MIT License\footnote{\hyperlink{https://opensource.org/licenses/MIT}{https://opensource.org/licenses/MIT}} - \texttt{amrlib}, \texttt{spacy}, \texttt{fairseq}, \texttt{smatch}, and \texttt{STRAP}. The following packages use the Apache License 2.0\footnote{\hyperlink{https://www.apache.org/licenses/LICENSE-2.0}{https://www.apache.org/licenses/LICENSE-2.0}} - \texttt{nltk}, and \texttt{huggingface's transformer}. The following packages use the GNU LGPL license 3.0\footnote{\hyperlink{https://www.gnu.org/licenses/lgpl-3.0.en.html}{https://www.gnu.org/licenses/lgpl-3.0.en.html}} - \texttt{gensim}.

\section{T-STAR-\textit{Encoder} Ablation Study}
\label{encoderablation}
The performance of our T-STAR-Encoder heavily depends on the quality of synthetic dataset generated while stylizing the sentences present in AMR 3.0 dataset. Therefore, we conducted thorough empirical analysis to identify a filtering strategy to boost the performance of the encoder. 

To obtain the initial set of stylized sentences, we use the state-of-the-art model available to transform generic English sentences to relevant styles \textit\{{Poetry, Shakespeare, Bible, Switchboard\}}, i.e., seq2seq inverse paraphrase module \cite{divpara}.

We then fine-tune our T-STAR-Encoder on synthetic datasets obtained from different filtering strategies, and compare the performan on test split of AMR 3.0, to ensure that the quality of the generated AMRs do not drop significantly. We present our findings in Table \ref{tab:text2amr-gold}. We observe that using the whole set of generated samples, leads to a significant drop in the performance (row-1). 

Therefore, we filter out the augmented stylized sentences with SIM similarity score \cite{sim} below the threshold $\delta$. This filtering strategy was able to significantly improve over the T-STAR-Encoder performance, giving competitive results to the non-augmented Vanilla T-STAR-Encoder. We select the best performing $\delta$ based on the performance on the test-split of AMR 3.0. Note that we have used the best performing threshold, $\delta=0.7$ 


\begin{table}[h]
\centering
\resizebox{\columnwidth}{!}{%
\begin{tabular}{|l|l|l|l|}
\hline
\textbf{Datasets} & 
\textbf{Precition} & \textbf{Recall} & \textbf{F-Score} \\ \hline
Vanilla T-STAR Encoder & 0.829 & 0.794 & 0.811 \\
T-STAR-Encoder & 0.671 & 0.390 & 0.493 \\
T-STAR Encoder-Flt ($\delta$=0.5) & 0.830 & 0.790 & 0.810 \\
T-STAR Encoder-Flt ($\delta$=0.6) & 0.807 & 0.753 & 0.779 \\
T-STAR Encoder-Flt ($\delta$=0.7) & 0.836 & 0.798 & 0.816 \\
T-STAR Encoder-Flt ($\delta$=0.8) & 0.828 & 0.793 & 0.810 \\
Iterative T-STAR-Encoder($\delta$=0.7) & 0.829 & 0.794 & 0.811 \\

\hline
\end{tabular}}
\caption{SMATCH scores on various versions of T-STAR-Encoder on AMR 3.0 dataset's test split.}
\label{tab:text2amr-gold}
\end{table}

\section{Unsupervised Evaluation of AMR parsing}
\label{app:wmd}
Since we use AMR graphs as the intermediate representation for the TST task, it is important to validate the generation quality of generated AMRs in terms of content preservation with respect to the input sentence. However, there does not exist an unsupervised metric to evaluate content overlap between an AMR graph and a sentence. Hence we propose to use a slight variation of the Word Mover Distance \cite{wmd} for this purpose. We choose WMD over other content preservation metrics for the following reasons - 
\begin{itemize}
    \item \citet{yamshchikov-etal-2019-dyr} illustrate the efficacy of WMD to evaluate text style transfer over other metrics based on correlation with human evaluations. Which means that it is more robust to the domain difference in the input and the output sentence, making it an ideal candidate for text-AMR similarity measurement.
    \item Syntactic metrics like BLEU would not be able to compute the content overlap between an AMR graph and a sentence because word representation in an AMR graph discards noun forms and tense information, and some verb tokens are mapped to a different PropBank verb. These modifications along with the disparity in sequential-graphical representation makes syntactic metrics infeasible for the task.
    \item Semantic representations like SIM are fragile to the input sequence order, and affected by non-content bearing words as well. However, WMD adopts on a bag-of-words paradigm, making it more suitable for the task.
\end{itemize}

We propose the following two variants of WMD - 
\begin{itemize}
    \item \textbf{WMD Overall} - In this variant, we aimed to keep the content bearing tokens from the sentence and AMR graphs. For sentence, we removed the stopwords (after doing a detailed corruption study on sentences, refer to Table \ref{tab:wmd-corruption}), while for AMR Graphs we removed AMR notation specific tokens (like ``:op?", ``ARG?"), punctuations (like ``(", `" '), assigned variables and propbank code for verbs (eg. changing ``s / say-01" to ``say").
    \item \textbf{WMD Verb Overall} - In this variant we specifically want to compute the similarity of verbs in the parsed AMR graphs and input sentence. For this, use \texttt{nltk} POS tagging tool to extract out verbs from the input sentence, and directly extract out propbank based verbs from the AMR graph.
\end{itemize}
Refer to Table \ref{tab:wmd-preprocessing} for an example of the preprocessing strategy adopted.

\begin{table}[h]
\centering
\resizebox{\columnwidth}{!}{%
\begin{tabular}{|l|}
\hline
    Input Sentence - \\
    \textit{Malaysian vice-prime minister Anwar ended a visit to China this} \\
    \textit{afternoon , and left Shanghai for Tokyo.} \\\hline
    Extracted content from Input Sentence - \\
    \textit{malaysian vice-prime minister anwar ended visit china afternoon ,} \\
    \textit{left shanghai tokyo.} \\\hline
    Extracted Verbs from Input Sentence - \\
    \textit{ended left} \\\hline
    Input AMR- \\
    \textit{(a2 / and
      :op1 (e2 / end-01
            :ARG0 (p / person} \\
                  \textit{:name (n / name
                        :op1 "Anwar")
                  :ARG0-of (h / have-org-role- 91} \\
                        \textit{:ARG1 (c7 / country
                              :name (n3 / name} \\
                                    \textit{:op1 "Malaysia"))
                        :ARG2 (m / minister} \\
                              \textit{:mod (p2 / prime)
                              :mod (v /} \\ 
                              \textit{vice))))
            :ARG1 (v2 / visit-01
                  :ARG0 p} \\
                  \textit{:ARG1 (c6 / country
                        :name (n2 / name
                              :op1} \\ \textit{"China")))
            :time (d / date-entity
                  :dayperiod (a3 / afternoon)} \\
                  \textit{:mod (t / today)))
      :op2 (l / leave-11} \\
            \textit{:ARG0 p
            :ARG1 (c8 / city
                  :name (n4 / name} \\
                        \textit{:op1 "Shanghai"))
            :ARG2 (c9 / city} \\
                  \textit{:name (n5 / name
                        :op1 "Tokyo"))))} \\\hline
    Extracted sequence- \\
        \textit{and end person name Anwar have-org-role country name} \\
        \textit{Malaysia minister prime vice visit country name China} \\
        \textit{afternoon today leave city name Shanghai city name Tokyo} \\\hline
    Corresponding Verb extraction (AMR) - \\
        \textit{end visit leave} \\\hline
\end{tabular}}
\caption{Illustrative example of the preprocessing done in proposed text-AMR unsupervised WMD Overall and WMD Verb Overall metrics.} 
\label{tab:wmd-preprocessing}
\end{table}

\begin{table}[h]
\centering
\resizebox{\columnwidth}{!}{%
\begin{tabular}{l|l|l}
\hline
\textbf{Model} & 
\textbf{WMD Mean / Std dev.}  & \textbf{SIM Mean / Std dev.}\\
\hline
Original & 0.0 / 0.0    &   1.0 / 0.0\\
Stop Words & 0.1049 / 0.0761    &   0.9373 / 0.0734 \\
Lowercase & 0.0754 / 0.1319     &   0.9946 / 0.0321 \\
Stop Words + Lowercase & 0.1663 / 0.1347    &   0.9373 / 0.0734 \\
POS & 0.1081 / 0.0731   &   0.8680 / 0.1929 \\ 
Stop Words + Lowercase + POS & 0.1966 / 0.1438  &   0.8492 / 0.1957 \\
Synonym Replacement & 0.1864 / 0.1266   &   0.6946 / 0.1832 \\
Synonym + Stop Words & 0.2228 / 0.1335  &   0.6545 / 0.1979 \\
Synonym + Lowercase & 0.2133 / 0.1511 & 0.6946 / 0.1832 \\
Synonym + Stop Words + Lowercase & 0.2476 / 0.1534  &   0.6545 / 0.1979 \\
Synonym + POS & 0.2320 / 0.1306     &   0.5948 / 0.2209 \\
Synonym + Stop Words + Lowercase + POS & 0.2686 / 0.1525    & 0.5817 / 0.2243 \\
\hline
\end{tabular}
}
\caption{WMD scores on various corruption strategies against original sentence on test set of AMR 3.0 dataset. POS refers to removing tags other than nouns, verbs and adjectives from the sentence.}
\label{tab:wmd-corruption}
\end{table}

\begin{figure}[h]
    \centering
    \includegraphics[width=0.5\textwidth]{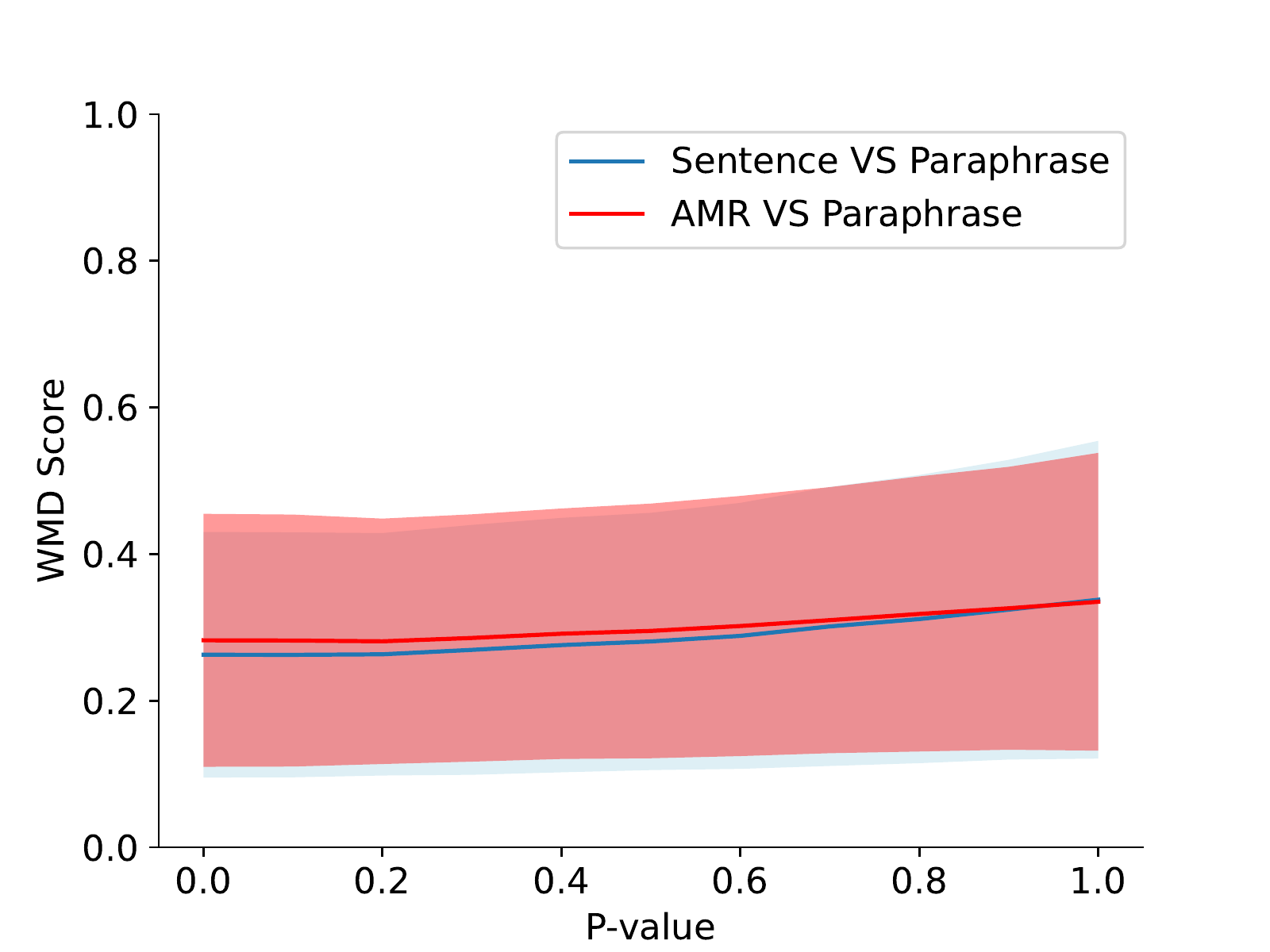}
    \caption{WMD Scores on AMR 3.0 dataset for sentence-paraphrase pairs (blue) and amr-paraphrase pairs (red). The shaded region denotes standard deviations from the mean.}
    \label{fig:wmd_metric}
\end{figure}

We also study the effect on text-AMR WMD score juxtaposed to text-text WMD scores on varying degree of similarity between the compared sequences. For this we use the diverse paraphraser trained by \citet{divpara} on the test set of AMR 3.0 dataset, and generate paraphrases with varying nucleus sampling p-values \cite{nucleus-sampling} from 0.0 to 1.0 with step size of 0.1. We notice that the WMD scores for text-text WMD (blue line in Fig. \ref{fig:wmd_metric}) and text-AMR WMD (red line in Fig. \ref{fig:wmd_metric}) are similar to each other throughout. 

We present the WMD Overall scores and WMD Verb Overall scores for the CDS dataset in Table \ref{tab:wmd-cds-scores}; validating the content retention in parsed AMR graphs for different strategies. We notice that Iterative T-STAR Encoder outperforms the T-STAR Encoder-Flt ($\delta$=0.7) and T-STAR Encoder (unfiltered) baselines in both WMD overall and Verb WMD overall. Even though comparable, it is still however lesser than the Vanilla T-STAR Encoder numbers. However, we believe we can credit that to more style information retention in Vanilla T-STAR Encoder (refer to Section \ref{sec:style-agnostic}), leading to poorer performance in downstream text style transfer task.

\begin{table}[]
\resizebox{\columnwidth}{!}{%
\begin{tabular}{|l|l|l|l|}
\hline
\textbf{Models}                                  & \textbf{Styles}      & \textbf{WMD Overall} & \textbf{Verb WMD Overall} \\
\hline
\multirow{4}{*}{Vanilla T-STAR Encoder}               & Bible       & 0.243       & 0.409            \\
                                        & Poetry      & 0.272       & 0.532            \\
                                        & Shakespeare & 0.322       & 0.484            \\
                                        & Switchboard & 0.292       & 0.467            \\\hline
\multirow{4}{*}{T-STAR Encoder (Unfiltered)}          & Bible       & 0.339       & 0.511            \\
                                        & Poetry      & 0.357       & 0.599            \\
                                        & Shakespeare & 0.403       & 0.552            \\
                                        & Switchboard & 0.341           & 0.537                \\\hline
\multirow{4}{*}{T-STAR Encoder Flt ($\delta$=0.7)} & Bible       & 0.290        & 0.461            \\
                                        & Poetry      & 0.322       & 0.568            \\
                                        & Shakespeare & 0.365       & 0.522            \\
                                        & Switchboard & 0.323       & 0.512    \\\hline
\multirow{4}{*}{Iterative T-STAR Encoder Flt ($\delta$=0.7)} & Bible       & 0.281        & 0.439            \\
                                        & Poetry      & 0.300       & 0.550            \\
                                        & Shakespeare & 0.344       & 0.500            \\
                                        & Switchboard & 0.301       & 0.491    \\
\hline
\end{tabular}}
\caption{WMD Overall and WMD Verb Overall scores for evaluation of unsupervised content preservation of different models across different styles in the CDS dataset \cite{divpara}.} \label{tab:wmd-cds-scores}
\end{table}

\section{Data Augmentation for T-STAR Decoder}
\label{app:wiki}
We hypothesize that the T-STAR decoder performances will improve if the underlying model, is better at generating text given an AMR graph. To this end, we create synthetic dataset using sentences from Wikipedia corpus. We sample 10 million sentences from it and generate the corresponding AMRs using our vanilla T-STAR model. We further filter out the samples for which WMD Overall, mentioned in Appendix \ref{app:wmd} and keep samples with a WMD score below 0.15, which results in 2.3M instances. 

We first fine-tune the T5-Base model for AMR to Text task on this filter dataset. We obtain a BLEU score of 49.13 on Gold AMR test set. Note that this performance is very close to the state-of-the art result 49.2 BLEU for this task \cite{bai2022graph}. Table \ref{wiki-models} lists down the different filtering strategy and dataset sizes we experimented to identify the best strategy to improve the performance.

\begin{table}[]
\small
\begin{tabular}{|l|r|}
\hline
\textbf{Dataset}                & \multicolumn{1}{l|}{\textbf{BLEU Score}} \\ \hline
GoldAMR (Baseline)              & 44.36                                    \\ \hline
Wikipedia 1M (Unfiltered)       & 48.14                                    \\ \hline
Wikipedia 120K- Filter($<$0.15)   & 44.56                                    \\ \hline
Wikipedia 1M -Filter ($<$0.15)    & 48.58                                    \\ \hline
Wikipedia 2.3M - Filter ($<$0.15) & \textbf{49.14}                           \\ \hline
\end{tabular}
\caption{Fine-tuning model for AMR-To-Text generation task using data augmentation technique using Wikipedia Sentences}
\label{wiki-models}
\end{table}

We then compare the performance of the best performing model on the style transfer task again STRAP. We observe that this model is not beating the STRAP performance across various style directions. Therefore, we conclude that a vanilla fine-tuning of model for AMR to text task, does not necessarily boost the performance in the downstream tasks. 

\section{Error Analysis}
\label{app:erroranalysis}
\subsection{Comparison on Meaning Preservation}
We present the results across all the 12 directions for content preservation comparitive analysis in Table \ref{human-eval-cp} and \ref{humaneval-cp-ittstar} respectively. We can observe that for every direction our models are consistently better in content preservation with respect to STRAP model. 

\begin{table}[h]
\resizebox{1\columnwidth}{!}{
\begin{tabular}{|llll|}
\hline
Direction                             & TSTAR \textgreater STRAP & STRAP \textless TSTAR & TSTAR=STRAP \\
\hline
bible $\rightarrow$ poetry            & 60.8\%                      & 37.6\%                   & 1.6\%           \\
bible $\rightarrow$ shak.       & 66.6\%                      & 32.2\%                   & 1.2\%           \\
bible$\rightarrow$switch.       & 67\%                      & 30.4\%                   & 2.6\%          \\
\hline
poetry$\rightarrow$bible            & 68.6\%                      & 29.4\%                   & 2\%          \\
poetry $\rightarrow$ shak.      & 78.6\%                      & 19.6\%                    & 1.8\%           \\
poetry$\rightarrow$switch.      & 69.4\%                      & 27.4\%                   & 3.2\%          \\
\hline
shak.$\rightarrow$bible       & 73\%                      & 24.8\%                   & 2.2\%          \\
shak.$\rightarrow$poetry      & 66.2\%                      & 27.4\%                   & 3.2\%          \\
shak.$\rightarrow$switch & 73.6\%                     & 24\%                   & 2.4\%          \\
\hline
switch.$\rightarrow$bible       & 73\%                      & 24\%                   & 3\%          \\
switch.$\rightarrow$poetry      & 73.6\%                      & 23.4\%                   & 3\%         \\
switch.$\rightarrow$shak. & 79.4\%                      & 18.6\%                    & 4\%       \\  
\hline
\end{tabular}}
\caption{Comparitive analysis of TSTAR and STRAP model to understand which model generates more meaning preserving outputs}
\label{human-eval-cp}
\end{table}

\begin{table}[h]
\resizebox{1\columnwidth}{!}{
\begin{tabular}{|llll|}
\hline
\textbf{Direction}                                  & \multicolumn{1}{l|}{\textbf{TSTAR=STRAP}} & \multicolumn{1}{l|}{\textbf{TSTAR $<$ STRAP}} & \multicolumn{1}{l|}{\textbf{TSTAR $>$ STRAP}} \\ \hline
\textit{bible$\rightarrow$switch.}                          & 1.8                                       & 31.6                                       & 66.6                                        \\ \hline
\textit{poetry$\rightarrow$switch.}                         & 2                                         & 15.6                                       & 82.4                                        \\ \hline
\textit{shak.$\rightarrow$switch.}                    & 4                                         & 20.6                                       & 75.4                                        \\  \hline
\textit{switch.$\rightarrow$bible}                          & 1.4                                       & 19.4                                       & 79.2                                        \\ \hline
\textit{switch.$\rightarrow$poetry}                         & 2.2                                       & 16.2                                       & 81.6                                        \\ \hline
\textit{switch.$\rightarrow$shak.}                    & 0.8                                       & 13                                         & 86.2                                        \\ \hline
\textit{bible$\rightarrow$poetry}       & 1.8                                       & 27.8                                       & 70.4                                        \\ \hline
\textit{bible$\rightarrow$shak.}  & 0.8                                       & 28.8                                       & 70.4                                        \\ \hline
\textit{poetry$\rightarrow$bible}       & 2                                         & 20.2                                       & 77.8                                        \\ \hline
\textit{poetry$\rightarrow$shak.} & 2.8                                       & 12.2                                       & 85                                          \\ \hline
\textit{shak.$\rightarrow$bible}  & 2.4                                       & 19.6                                       & 78                                          \\ \hline
\end{tabular}}
\caption{Comparison on content preservation using human evaluations on Iterative T-STAR and STRAP across all 12 direction}
\label{humaneval-cp-ittstar}
\end{table}

\begin{table*}[h]
\small
\resizebox{1\textwidth}{!}{
\begin{tabular}{|l|rrr|rrr|rrr|rrr|}
\hline
& STRAP & TSTAR & Itr-TSTAR   &STRAP & TSTAR & Itr-TSTAR                      &STRAP & TSTAR & Itr-TSTAR                      &STRAP & TSTAR & Itr-TSTAR                   \\ \cline{2-13} 
\multicolumn{1}{|l|}{\multirow{-2}{*}{\textbf{Direction}}} & \multicolumn{3}{c|}{No Error $\uparrow$}                                                                              & \multicolumn{3}{c|}{Hallucination $\downarrow$}                                                                         & \multicolumn{3}{c|}{Incomplete $\downarrow$}                                                                            & \multicolumn{3}{c|}{Semantic Drift $\downarrow$}                                                                        \\ \hline
\multicolumn{1}{|l|}{bible$\rightarrow$poetry}                & \multicolumn{1}{r|}{42}    & \multicolumn{1}{r|}{43}    & \multicolumn{1}{r|}{62}  & \multicolumn{1}{r|}{98}    & \multicolumn{1}{r|}{68}    & \multicolumn{1}{r|}{70}  & \multicolumn{1}{r|}{239}   & \multicolumn{1}{r|}{270}   & \multicolumn{1}{r|}{228} & \multicolumn{1}{r|}{121}   & \multicolumn{1}{r|}{119}   & \multicolumn{1}{r|}{140} \\ \hline
\multicolumn{1}{|l|}{bible$\rightarrow$shakespeare}           & \multicolumn{1}{r|}{45}    & \multicolumn{1}{r|}{72}    & \multicolumn{1}{r|}{93}  & \multicolumn{1}{r|}{224}   & \multicolumn{1}{r|}{76}    & \multicolumn{1}{r|}{87}  & \multicolumn{1}{r|}{55}    & \multicolumn{1}{r|}{190}   & \multicolumn{1}{r|}{149} & \multicolumn{1}{r|}{176}   & \multicolumn{1}{r|}{162}   & \multicolumn{1}{r|}{171} \\ \hline
\multicolumn{1}{|l|}{bible$\rightarrow$switchboard}           & \multicolumn{1}{r|}{54}    & \multicolumn{1}{r|}{98}    & \multicolumn{1}{r|}{106} & \multicolumn{1}{r|}{203}   & \multicolumn{1}{r|}{65}    & \multicolumn{1}{r|}{87}  & \multicolumn{1}{r|}{59}    & \multicolumn{1}{r|}{189}   & \multicolumn{1}{r|}{132} & \multicolumn{1}{r|}{184}   & \multicolumn{1}{r|}{148}   & \multicolumn{1}{r|}{175} \\ \hline
\multicolumn{1}{|l|}{poetry$\rightarrow$bible}                & \multicolumn{1}{r|}{46}    & \multicolumn{1}{r|}{85}    & \multicolumn{1}{r|}{110} & \multicolumn{1}{r|}{257}   & \multicolumn{1}{r|}{232}   & \multicolumn{1}{r|}{175} & \multicolumn{1}{r|}{22}    & \multicolumn{1}{r|}{27}    & \multicolumn{1}{r|}{15}  & \multicolumn{1}{r|}{175}   & \multicolumn{1}{r|}{156}   & \multicolumn{1}{r|}{200} \\ \hline
\multicolumn{1}{|l|}{poetry$\rightarrow$shakespeare}          & \multicolumn{1}{r|}{43}    & \multicolumn{1}{r|}{134}   & \multicolumn{1}{r|}{166} & \multicolumn{1}{r|}{224}   & \multicolumn{1}{r|}{89}    & \multicolumn{1}{r|}{79}  & \multicolumn{1}{r|}{50}    & \multicolumn{1}{r|}{53}    & \multicolumn{1}{r|}{47}  & \multicolumn{1}{r|}{183}   & \multicolumn{1}{r|}{224}   & \multicolumn{1}{r|}{208} \\ \hline
\multicolumn{1}{|l|}{poetry$\rightarrow$switchboard}          & \multicolumn{1}{r|}{80}    & \multicolumn{1}{r|}{142}   & \multicolumn{1}{r|}{160} & \multicolumn{1}{r|}{171}   & \multicolumn{1}{r|}{63}    & \multicolumn{1}{r|}{39}  & \multicolumn{1}{r|}{63}    & \multicolumn{1}{r|}{82}    & \multicolumn{1}{r|}{65}  & \multicolumn{1}{r|}{186}   & \multicolumn{1}{r|}{213}   & \multicolumn{1}{r|}{236} \\ \hline
\multicolumn{1}{|l|}{shakespeare$\rightarrow$bible}           & \multicolumn{1}{r|}{44}    & \multicolumn{1}{r|}{69}    & \multicolumn{1}{r|}{120} & \multicolumn{1}{r|}{253}   & \multicolumn{1}{r|}{203}   & \multicolumn{1}{r|}{186} & \multicolumn{1}{r|}{19}    & \multicolumn{1}{r|}{43}    & \multicolumn{1}{r|}{14}  & \multicolumn{1}{r|}{184}   & \multicolumn{1}{r|}{185}   & \multicolumn{1}{r|}{180} \\ \hline
\multicolumn{1}{|l|}{shakespeare$\rightarrow$poetry}          & \multicolumn{1}{r|}{44}    & \multicolumn{1}{r|}{55}    & \multicolumn{1}{r|}{85}  & \multicolumn{1}{r|}{187}   & \multicolumn{1}{r|}{190}   & \multicolumn{1}{r|}{161} & \multicolumn{1}{r|}{77}    & \multicolumn{1}{r|}{89}    & \multicolumn{1}{r|}{69}  & \multicolumn{1}{r|}{192}   & \multicolumn{1}{r|}{166}   & \multicolumn{1}{r|}{187} \\ \hline
\multicolumn{1}{|l|}{shakespeare$\rightarrow$switchboard}     & \multicolumn{1}{r|}{70}    & \multicolumn{1}{r|}{132}   & \multicolumn{1}{r|}{161} & \multicolumn{1}{r|}{151}   & \multicolumn{1}{r|}{69}    & \multicolumn{1}{r|}{62}  & \multicolumn{1}{r|}{42}    & \multicolumn{1}{r|}{98}    & \multicolumn{1}{r|}{47}  & \multicolumn{1}{r|}{237}   & \multicolumn{1}{r|}{201}   & \multicolumn{1}{r|}{230} \\ \hline
\multicolumn{1}{|l|}{switchboard$\rightarrow$bible}           & \multicolumn{1}{r|}{50}    & \multicolumn{1}{r|}{85}    & \multicolumn{1}{r|}{129} & \multicolumn{1}{r|}{241}   & \multicolumn{1}{r|}{213}   & \multicolumn{1}{r|}{221} & \multicolumn{1}{r|}{42}    & \multicolumn{1}{r|}{54}    & \multicolumn{1}{r|}{14}  & \multicolumn{1}{r|}{167}   & \multicolumn{1}{r|}{148}   & \multicolumn{1}{r|}{135} \\ \hline
switchboard$\rightarrow$poetry                      & 
\multicolumn{1}{r|}{49}               & 
\multicolumn{1}{r|}{98}               & 
\multicolumn{1}{r|}{114}            & 
\multicolumn{1}{r|}{165}              & 
\multicolumn{1}{r|}{113}              & 
\multicolumn{1}{r|}{97}             & \multicolumn{1}{r|}{140}& 
\multicolumn{1}{r|}{165}              & 
\multicolumn{1}{r|}{123}            & 
\multicolumn{1}{r|}{146}              & 
\multicolumn{1}{r|}{124}              & 
\multicolumn{1}{r|}{166}            \\ \hline
switchboard$\rightarrow$shakespeare                 & 
\multicolumn{1}{r|}{62}               & 
\multicolumn{1}{r|}{149}              & 
\multicolumn{1}{r|}{186}            & 
\multicolumn{1}{r|}{189}              & 
\multicolumn{1}{r|}{87}               & 
\multicolumn{1}{r|}{92}             & 
\multicolumn{1}{r|}{77}               & 
\multicolumn{1}{r|}{87}               & 
\multicolumn{1}{r|}{31}             & 
\multicolumn{1}{r|}{172}              & 
\multicolumn{1}{r|}{177}              & 
\multicolumn{1}{r|}{191} \\ \hline         
\end{tabular}}
\caption{Type of Error Analysis across three models STRAP, TSTAR, Iterative-TSTAR across all the four styles.}
\label{tab:error-analysis}
\end{table*}

\subsection{Error Analysis per direction}
In this section, we present the error analysis for each direction in Table~\ref{tab:error-analysis}. We observe that T-STAR and Iterative T-STAR models are consistently better on No-Error and Hallucinations across all the directions. Moreover, we observe that for 7 out of 12 direction, Iterative T-STAR model is better than STRAP for Incompleteness error. Note that our T-STAR model was under-performing, however another iteration of model improvements increases that number significantly. Note that all the models are giving high error in Semantic Drift, and improving the model for this type of errors can be explored in future works.

\begin{table*}[]
\small
\centering
\resizebox{0.8\textwidth}{!}{
\begin{tabular}{|l|p{5cm}|p{5cm}|l|}
\hline\hline
\textbf{S.No} & \textbf{Input Stylized Sentence}                          & \textbf{Output Stylized Sentence}                                                                      & \textbf{TSTAR-Encoder AMR}                                                                                                                                                                                                                                                                                                                                                                                                                                                                                                                                                                                                                                                                                                                                                                                                                                                                                                                                                                                                                         \\\hline
1. & And she answered and said unto him, Yes, Lord: yet the dogs under the table eat of the children's crumbs.                                                                            & Yet the dogs under the table Eat children’s crumbs.                                           & \begin{tabular}[c]{@{}l@{}}(e / eat-01\\\quad:ARG0 (d / dog\\\quad\quad:location (u / under\\\quad\quad\quad:op1 (t / table)))\\\quad:ARG1 (c / crumb\\\quad\quad:poss (c2 / child))\\\quad:mod (s / still))\end{tabular}                                                                                                                                                                                                                                                                                                                                                                                                                                                                                                                                                                                                                                                        \\ \hline
2. & And the priest said unto them, Go in peace: before the LORD is your way wherein ye go.                                                                                               & Before the Lord go in peace on high;                                                          & \begin{tabular}[c]{@{}l@{}}(g / go-02\\\quad:mode imperative\\\quad:ARG0 (y / you)\\\quad:manner (p / peace)\\\quad:time (b / before\\\quad\quad:op1 (p2 / person\\\quad\quad\quad:name (n / name\\\quad\quad\quad\quad:op1 "Lord"))))\end{tabular}                                                                                                                                                                                                                                                                                                                                                                                                                                                                                                                                                                                \\ \hline
3. & And there were made on them, on the doors of the temple, cherubims and palm trees, like as were made upon the walls; and there were thick planks upon the face of the porch without. & The cherubims and palm-trees are at the temple doors, And the thick planks on the porch face. & \begin{tabular}[c]{@{}l@{}}(a / and\\\quad:op1 (b / be-located-at-91\\\quad\quad:ARG1 (a2 / and\\\quad\quad\quad:op1 (c / cherubim)\\\quad\quad\quad:op2 (t / tree\\\quad\quad\quad\quad:mod (p / palm)))\\\quad\quad:ARG2 (d / door\\\quad\quad\quad:part-of (t2 / temple)))\\\quad:op2 (b2 / be-located-at-91\\\quad\quad:ARG1 (p2 / plank\\\quad\quad\quad:ARG1-of (t3 / thick-03))\\\quad\quad:ARG2 (f / face\\\quad\quad\quad:part-of (p3 / porch))))\end{tabular}                                                                                                       \\ \hline
4. & and uh they've had catalytic con- you know catalytic converters on those on uh i think they're required in California and Oregon and and Washington                                  & In Caledonia, Oregon, and Washington catalytic conversion required.                           & \begin{tabular}[c]{@{}l@{}}(r / require-01\\\quad:ARG1 (t / thing\\\quad\quad:ARG0-of (c / convert-01\\\quad\quad\quad:mod (c2 / catalytic)))\\\quad:location (a / and\\\quad\quad:op1 (s / state\\\quad\quad\quad:name (n / name\\\quad\quad\quad\quad:op1 "California"))\\\quad\quad:op2 (s2 / state\\\quad\quad\quad:name (n2 / name\\\quad\quad\quad\quad:op1 "Oregon"))\\\quad\quad:op3 (s3 / state\\\quad\quad\quad:name (n3 / name\\\quad\quad\quad\quad:op1 "Washington"))))\end{tabular} \\ \hline
5.& That crowns a lofty clift, which proudly towers                                                                                                                                      & crowns the lofty towers                                                                       & \begin{tabular}[c]{@{}l@{}}(c / crown-01\\\quad :ARG0 (ii / it)\\ \quad:ARG1 (t / tower\\\quad\quad:mod (l / lofty)))\end{tabular}                                                                                                                                                                                                                                                                                                                                                                                                                                                                                                                                                                                                                                                                                                                                                                                                                                              \\ \hline
6. & And the rough goat is the king of Grecia: and the great horn that is between his eyes is the first king.                                                                             & The rough goat o'er the king of Greece                                                        & \begin{tabular}[c]{@{}l@{}}(h / have-org-role-91\\\quad:ARG0 (g / goat\\\quad\quad:ARG1-of (r / rough-04))\\\quad:ARG1 (c / country\\\quad\quad:name (n / name\\\quad\quad\quad:op1 "Greece"))\\\quad:ARG2 (k / king))\end{tabular}     \\                                                                                                      \hline                                                                                                                                                                                                                                                                                                                                                                                                                                                                                                    
\end{tabular}}
\caption{Samples from Human Evaluations where the T-STAR output was marked incomplete. We observe that entities and relations missing in the sentence, were not present in their corresponding intermediate AMRs.}
\label{amr-incomplete}
\end{table*}
\begin{table*}[h]
\small
\centering
\resizebox{0.8\textwidth}{!}{
\begin{tabular}{|p{5cm}|p{5cm}|l|l|}\hline\hline
Input Stylized Sentence                                                   & Output Stylized Sentence                                & Input generated AMR                                                                                                                                                                                                                                                                 & Potential Error \\\hline
Ay, mine own fortune in my misery.                                        & my fortune is my misery                                 & \begin{tabular}[c]{@{}l@{}}(m / misery\\ \quad:poss (ii / i)\\ \quad:domain (f / fortune\\ \quad\quad:poss ii))\end{tabular}                                                                                                                                                        & T-STAR Decoder     \\\hline
yeah it makes you wonder um had the jury been a different group of people & It maketh you to wonder at the diversities of the jury. & \begin{tabular}[c]{@{}l@{}}(m / make-02\\ \quad:ARG0 (ii / it)\\ \quad:ARG1 (w / wonder-01\\ \quad\quad:ARG0 (y / you)\\ \quad\quad:ARG1 (d / differ-02\\ \quad\quad\quad:ARG1 (j / jury))))\end{tabular}                                                                           & T-STAR Decoder     \\\hline
Tis not Maria's whispering call;                                          & It is not whispered by Maria.                           & \begin{tabular}[c]{@{}l@{}}(w / whisper-01\\ \quad:polarity -\\ \quad:ARG0 (p / person\\ \quad\quad:name (n / name\\ \quad\quad\quad:op1 "Maria"))\\ \quad:ARG1 (ii / it))\end{tabular}                                                                                             & T-STAR Encoder     \\\hline
Yea,” quoth he, “Dost thou fall upon thy face?                            & Quoth he, "Dost thou fall thy face,                     & \begin{tabular}[c]{@{}l@{}}(q / quote-01\\ \quad:ARG1 (f / fall-01\\ \quad\quad:ARG1 (y / you\\ \quad\quad\quad:part (f2 / face))\\ \quad\quad:ARG4 f2\\ \quad\quad:polarity (a / amr-unknown))\\ \quad:ARG2 (h / he))\end{tabular}                                                 & T-STAR Encoder     \\\hline
Around my grave they'll wither.                                           & They will wither about my grave.                        & \begin{tabular}[c]{@{}l@{}}(w / wither-01\\ \quad:ARG1 (t / they)\\ \quad:location (a / around\\ \quad\quad:op1 (g / grave\\ \quad\quad\quad:poss (ii / i))))\end{tabular}                                                                                                          & T-STAR Decoder     \\\hline
Justice is sworn 'gainst tears, and hers would crave                      & uh Justice has sworn to cried and cried                 & \begin{tabular}[c]{@{}l@{}}(s / swear-01\\ \quad:ARG0 (p / person\\ \quad\quad:name (n / name\\ \quad\quad\quad:op1 "Justice"))\\ \quad:ARG1 (a / and\\ \quad\quad:op1 (c / cry-02\\ \quad\quad\quad:ARG0 p)\\ \quad\quad:op2 (c2 / cry-02\\ \quad\quad\quad:ARG0 p)))\end{tabular} & T-STAR Encoder    \\\hline
\end{tabular}}
\caption{Various Instances that had Semantic Drift as a type of error. We manually analyze the same, and hypothesize that the potential error in the listed module.}
\label{amr-semantic}
\end{table*}
\subsection{Qualitative Analysis for Incompleteness and Semantic Drift}

As we are using interpretable intermediate representations, it is easily possible to understand the intuition behind these errors, and broadly understand which modules (encoder or decoder) needs to be improved further. Therefore, we study few instances and analyze the generated Intermediate AMRs to understand the reason for high number in \textit{Semantic Drift} and \textit{Incomplete} errors. We list down some intstances in Table \ref{amr-semantic} and Table \ref{amr-incomplete}

Across the various instances that we analyzed, we observe that the generated AMRs were not encoding the complete information themselves. For instance,  either missing some entities (example 1, 4, 5 and 6 in Table \ref{amr-incomplete}), if the clauses were separated using ":", ";", only one of those were parsed in the intermediate graph (example 2 and 3 in Table \ref{amr-incomplete}. For semantic drift, we observed that the errors was arising due to the  shortcomings in both the modules, i.e., it was leading to meaning change if the Encoder didn't generate an efficient AMR graph, or if the decoder was not able to interpret AMR correctly. We have listed the modules that could be the potential reason for the error in the last column in Table \ref{amr-semantic}.  

It is important to note that, the source of errors is very easy to identify now because we are using robust, interpretable and symbolic representation as pivot to transfer from style A to style B. We have also provided a case study of performance of various baselines and proposed model in Table \ref{tab:case-study} on the CDS dataset.

\end{document}